\documentclass[12pt]{report}

\usepackage[a-1b]{pdfx} 

\usepackage{pdfpages}

\newdimen \jot \jot=5mm
\usepackage[T1]{fontenc}
\usepackage{lmodern} 
\usepackage{tocbibind} 
\usepackage{setspace}
\usepackage{RJournal_nogeom} 
\usepackage[hypcap=true]{caption}
\hypersetup{linktocpage}
\usepackage{amsmath,amssymb,array}
\usepackage{booktabs}
\usepackage{subfig}
\usepackage{amssymb}

\usepackage{graphicx}
\usepackage{float}
\usepackage{tikz}
\usepackage{graphics} 
\usetikzlibrary{positioning}
\usetikzlibrary{shapes,arrows}
\usepackage{dcolumn}

\usepackage{color}
\makeatletter
\def\maxwidth{ %
  \ifdim\Gin@nat@width>\linewidth
    \linewidth
  \else
    \Gin@nat@width
  \fi
}
\makeatother

\definecolor{fgcolor}{rgb}{0.345, 0.345, 0.345}
\usepackage{framed}
\makeatletter
 {\par\unskip\endMakeFramed%
 \at@end@of@kframe}
\makeatother

\definecolor{shadecolor}{rgb}{.97, .97, .97}
\definecolor{messagecolor}{rgb}{0, 0, 0}
\definecolor{warningcolor}{rgb}{1, 0, 1}
\definecolor{errorcolor}{rgb}{1, 0, 0}
\newenvironment{knitrout}{}{} 
\makeatletter
\newcommand\gobblepars{%
    \@ifnextchar\par%
        {\expandafter\gobblepars\@gobble}%
        {}}
\makeatother

\usepackage[
style = authoryear, 
sorting = none, 
dashed = false,
maxbibnames = 99,
backend = bibtex,
natbib = true
]{biblatex}

\AtEveryBibitem{
\clearfield{note}
\clearfield{month}
}

\usepackage{enumerate}
\usepackage{amsthm} 

\graphicspath{{rnw_chapter/figure/}{rnw_chapter/}} 

\begin{document}

\newcommand{\bm}[1]{ \mbox{\boldmath $ #1 $} }
\newcommand{\bin}[2]{\left(\begin{array}{@{}c@{}} #1 \\ #2
             \end{array}\right) }
\renewcommand{\contentsname}{Table of Contents}
\baselineskip=24pt

\pagenumbering{roman}
\thispagestyle{empty}
\begin{center}
\vspace*{.25in}
{\bf\LARGE{ Contrastive Predictive Coding Based Feature for Automatic Speaker Verification }}\\ 
\vspace*{.75in}
{\bf by} \\*[18pt]
\vspace*{.2in}
{\bf Cheng-I Jeff Lai}\\ 
\vspace*{1in}
{\bf A thesis submitted to The Johns Hopkins University\\
in conformity with the requirements for the degree of\\
Bachelor of Science }\\
\vspace*{.75in}
{\bf Baltimore, Maryland} \\
{\bf December, 2018} \\     
\vspace*{.5in}
\begin{small}
{\bf \copyright{ }2018 by Cheng-I Jeff Lai} \\ 
{\bf All rights reserved}
\end{small}
\end{center}
\newpage 

\pagestyle{plain}
\pagenumbering{roman}
\setcounter{page}{2}
\chapter*{Abstract}
This thesis describes our ongoing work on Contrastive Predictive Coding (CPC) features for speaker verification. CPC is a recently proposed representation learning framework based on predictive coding and noise contrastive estimation. We focus on incorporating CPC features into the standard automatic speaker verification systems, and we present our methods, experiments, and analysis. This thesis also details necessary background knowledge in past and recent work on automatic speaker verification systems, conventional speech features, and the motivation and techniques behind CPC. 











\chapter*{Acknowledgments}

I have the privilege to be advised by one of the best in the field of speech processing and surrounded by talented and motivated individuals who encourages me to make strides for the speech community. I want to give my sincere gratitude to my advisor at JHU, Professor Najim Dehak, who introduced me to speech processing and pushed me to become better every day. There were times that I could not make good progress and felt like giving up, and it was Najim who supported me through those difficult moments. I could never forget the trust Najim has given to me and he is the best advisor I can ask for as an undergraduate. Thanks to Dr. Jes\'us Villalba, who has infinite patience for me in the past two years. I came in to the field with little knowledge in machine learning and coding, and it was Jes\'us who guided me step by step and taught me to be persistent in research. Thanks to Professor Simon King, who hosted me at University of Edinburgh and gave me the resources, guidance, and research environment to work on anti-spoofing. I have the best summer ever at Edinburgh without a doubt. Thanks to Professor Hynek Hermansky, who advised me the importance of the basics and scrutiny of conducting good research. Thansk to Professor Korin Richmond, Professor Junichi Jamagishi, and Professor Alberto Abad, who patiently answered several questions I have on anti-spoofing during the several meetings we had. I would also like to thank Laure moro, for the Parkinson's disease project and his help on improving my presentation skills, Phani Nidadavolu, for the bandwidth extension project and the good practices he taught me in conducting experiments, and Nanxin Chen, for helping me to learn coding, debug, discover sources for new research paper, and inspiring me to do creative work. 

I would like to extend my gratitude to other members in the CLSP group at JHU and the CSTR group at University of Edinburgh, especially Professor Daniel Povey, Professor Shinji Watanabe, Professor Alan Yuille, Professor Colin Wilson, Professor Mounya Elhilali, Paola Garcia-Perera, Dimitra Emmanouilidou, Debmalya Chakrabarty, Arun Nair, Matthew Wiesner, David Synder, Lucas Ondel, Aswin Subramanian, Ruizhi Li, ChuCheng Lin, Raghavendra Reddy, JaeJin Cho, Saurabh Bhati, Peter Frederiksen, Saurabh Kataria, Xutai Ma, Xiaofei Wang, Kelly Marchisio, Sray Chen, Cassia Valentini, Catherine Lai, Joanna R\'ownicka, Julie-Anne Meaney, Mark Sinclair, Felipe Espic, and Pacco.

I met a lot of brilliant people during college. Thanks to William Zhang, James Shamul, Justin Chua, Bijan Varjavand, Aurik Sarker, Harrison Nguyen, Eric Tsai, Kiki Chang, Vladimir Postnikov, Esther Tien, William Shyr, Jeff Peng, Kevin Chen, Kevin Ma, Chris Hong, Chin-Fu Liu, Ray Cheng, Tom Shen, Max Novick, Adriana Donis, Jillian Ho, Richard Oh, Alejandro Contreras, Cindy Yuan, Allen Ren, Linh Tran, Charlie Wang, Weiwei Lai, Michael Chan, and Emily Sun, for their love and support. Their kindness and presence have made all the differences in my life. I want to give special thanks to Daniel Hsu, who took good care of me when I was suffering from a herniated disc and sciatica, and for being an awesome roommate and friend. Finally, I want to thank my family members in Taiwan for supporting me emotionally and financially. I am especially grateful for my mom, who has always encouraged me to venture to a bigger world and do greater things.

\pagestyle{plain}
\baselineskip=24pt
\tableofcontents
\listoftables
\listoffigures

\cleardoublepage 
\pagenumbering{arabic}





\begin{refsection}[main.bib]
\chapter{Automatic Speaker Verification}
\label{chap:intro_speaker}

\section{Introduction}\footnote{The organization of this Chapter is inspired from Nanxin Chen's Center for Language and Speech Processing Seminar Talk "Advances in speech representation for speaker recognition".}
Speech is the main medium we use to communicate with the others, and therefore it contains rich information of our interests. Upon hearing a speech, in addition to identify what its content, it is natural for us to ask: Who is the speaker? What is the nationality of the speaker? What is his/her emotion?

Speaker Recognition is the collection of techniques to either identifies or verifies the speaker-related information of segments of speech utterances, and Automatic Speaker Recognition is speaker recognition performed by machines. Figure \ref{fig:speaker_detection_overview} is an overview of the speaker information in speech. Speaker information is embedding in speech, but it is often corrupted by channel effects to some degree. Channel effects can be environment noises, and more often recording noises since automatic speaker recognition is performed on speech recordings. There are some speaker-related information we are also interested in, such as age, emotion and language. 

This Chapter first gives a overview of Automatic Speaker Verification. Then several major speaker verification techniques, from the earlier Gaussian Mixture Models to the recent neural models, are presented subsequently.
    
\begin{knitrout}
\definecolor{shadecolor}{rgb}{0.969, 0.969, 0.969}\color{fgcolor}\begin{figure}
\centering\includegraphics[width=\maxwidth]{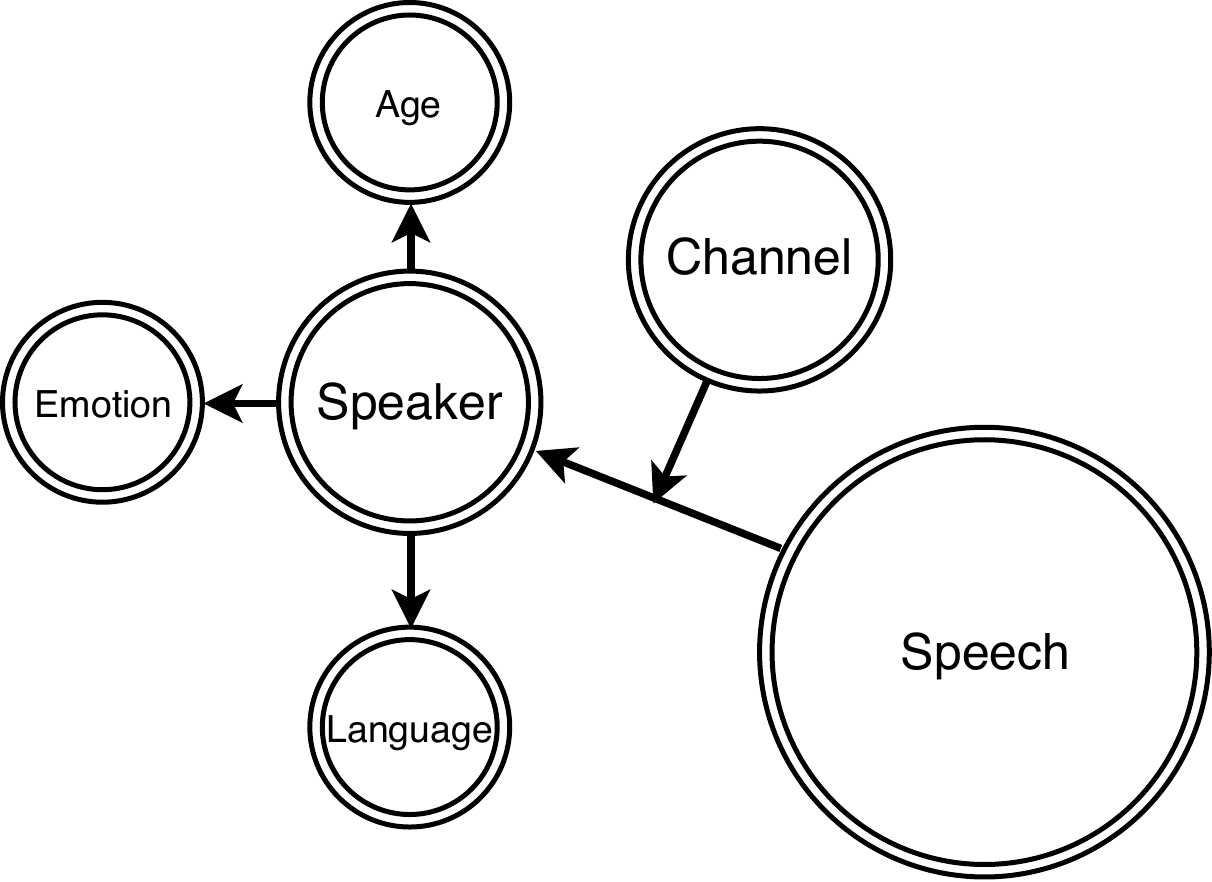} \caption[{\bf Speaker Detection: An Overview} ]{{\bf An overview of speaker information in speech.} Speaker information is embedded in speech and it is often disrupted by channel noises. From the speaker information, age, emotion, language, etc. of the speech can be inferred.}\label{fig:speaker_detection_overview}
\end{figure}
\end{knitrout}

    \subsection{Speaker Identification v.s. Verification}
    Speaker Recognition concerns with speaker-related information. Automatic Speaker Recognition is therefore the machines that perform speaker recognition for humans. Speaker Recognition can be categorized into Speaker Identification and Speaker Verification, by the testing protocol (Figure \ref{fig:identification_vs_verification}). As with any machine learning models, Automatic Speaker Recognition requires training data and testing data. Speaker Identification is to identify whether the speaker of a testing utterance matches any training utterances, and hence it is a closed-set problem. On the other hand, Speaker Verification is to verify weather the speakers of a pair of utterances match. The pair is consisted of an enrollment utterance and a testing utterance, which may not be presented beforehand, and hence it is a more challenging open-set problem.

    This thesis work focuses on Automatic Speaker Verification. 
    
    \begin{knitrout}
    \definecolor{shadecolor}{rgb}{0.969, 0.969, 0.969}\color{fgcolor}\begin{figure}
    \centering\includegraphics[width=0.75\maxwidth]{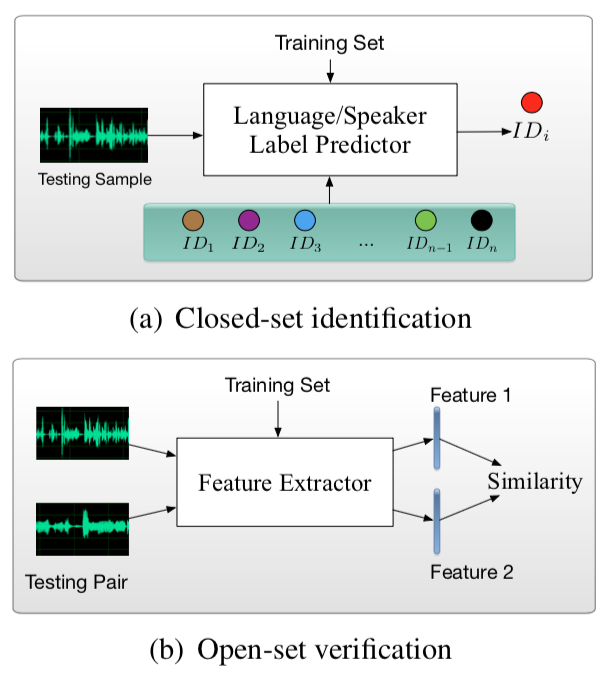} \caption[{\bf Speaker Identification versus Verification} ]{{\bf Speaker identification v.s. Speaker verification \citep{cai2018exploring}.} Speaker identification can be framed as a closed-set problem, while verification can be framed as an open-set problem.}\label{fig:identification_vs_verification}
    \end{figure}
    \end{knitrout}
    
    \subsection{General Processing Pipeline}
    Figure \ref{fig:sr_pipeline} describes the four main stages of Automatic Speaker Recognition (thus includes Verification). Most systems have these four aspects in their system design. Feature Processing is to get low-level feature descriptors from the speech waveforms, such as Mel-Frequency Cepstral Coefficients (MFCC), FilterBank, Perceptual Linear Predictive (PLP) Analysis, or bottleneck features. Clustering is the process to differentiate different acoustic units and process them separately, and it is commonly adopted in speaker recognition, such as Gaussian Mixture Model (GMM). Summarization is the conversion from variable-length frame-level features to a fixed-length utterance-level feature, such as the i-vectors or average pooling. Backend Processing is for scoring and making decisions, such as Support Vector Machine (SVM), Cosine Similarity or Probablistic Linear Discriminant Analysis (PLDA).

    \begin{knitrout}
    \definecolor{shadecolor}{rgb}{0.969, 0.969, 0.969}\color{fgcolor}\begin{figure}
    \centering\includegraphics[width=\maxwidth]{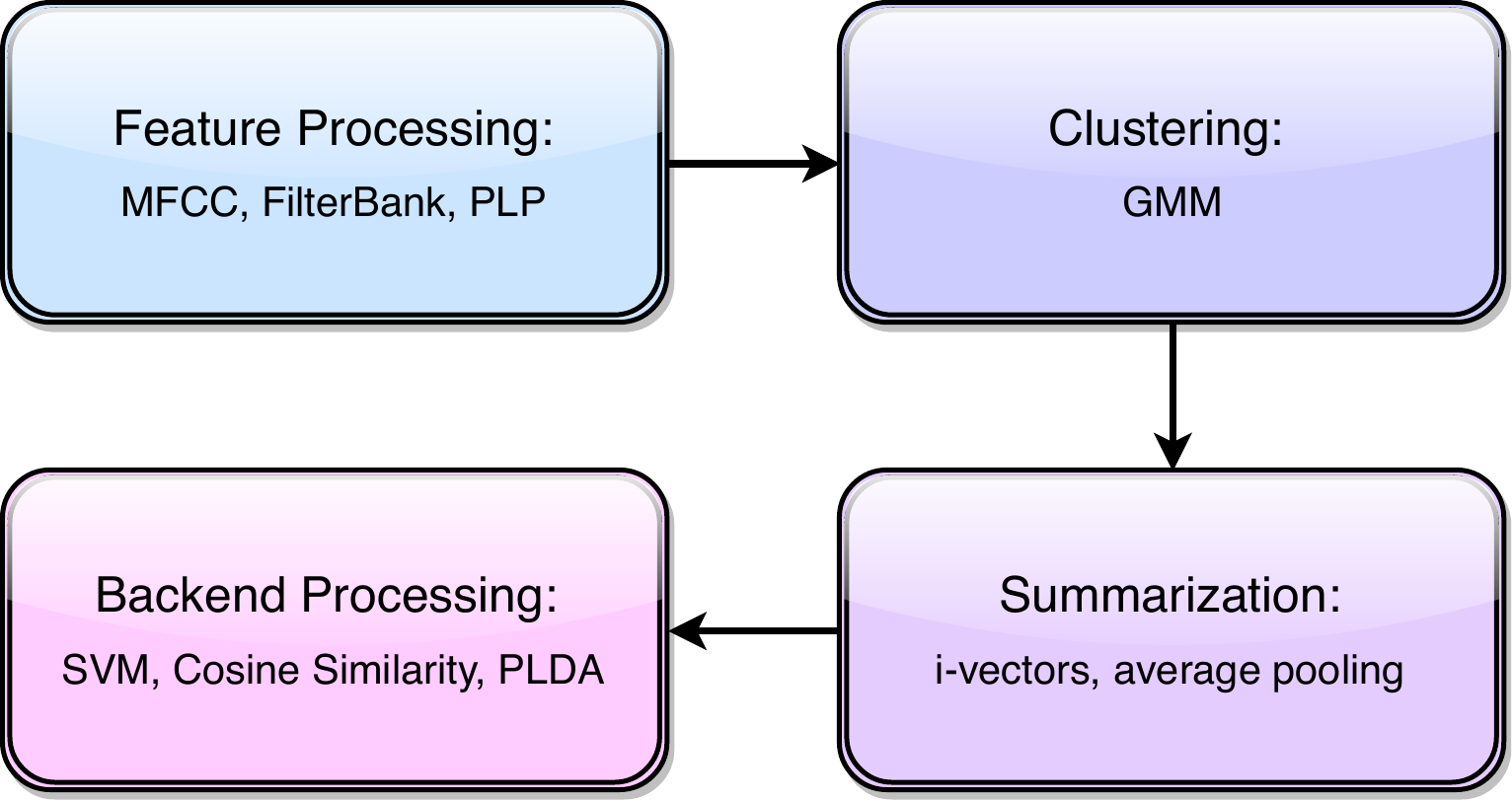} \caption[{\bf Speaker Recognition General Pipeline} ]{{\bf Four stages of speaker recognition: Feature Processing, Clustering, Summarization, and Backend Processing.} Feature Processing is to get low-level features from speech utterances, such as MFCC, FilterBank, and PLP. Clustering is the process to differentiate different acoustic units and process them separately, such as GMM. Summarization is the conversion from variable-length frame-level features to a fixed-length utterance-level feature, such as the i-vectors. Backend Processing is for scoring and making decisions, such as SVM, Cosine Similarity and PLDA. 
    }\label{fig:sr_pipeline}
    \end{figure}
    \end{knitrout}

    \subsection{Metrics}
    There are various metrics defining how well a system performs, such as the Decision Cost Functio (DCF) and Equal Error Rate (EER). DCF is defined as:
    \begin{flalign}
    C_{DET}(\theta) = C_{FRR}\times P_{Target}\times P_{FRR}(\theta) + C_{FAR}\times (1-P_{Target})\times P_{FAR}(\theta)
    \end{flalign}
    EER is the equilibrium point between False Alarm Rate and False Negative Rate. We adopt EER for this thesis work for its common use in Automatic Speaker Recognition work.
    
    \subsection{Challenges}
    Speaker Recognition at its core is to optimize a Sequence-to-One mapping function. From the task perspective, it is supposedly easier than Sequence-to-Sequence tasks since it only outputs once per sequence. However, from the data perspective, it is much harder. Comparing to automatic speech recognition or machine translation, which are Sequence-to-Sequence mappings, there is very little data for automatic speaker recognition. For example, a 100 seconds YouTube video could have more than 100 words spoken but only 1 speaker identity. In addition to data, channel effects have been the major bottleneck for previous research work on speaker recognition (Figure \ref{fig:speaker_detection_overview}). Advances in the field has developed techniques that aim to address it, such as the Joint Factor Analysis, but channel effects still play a significant role. This is one reason why the most fundamental task in speech, voice activity detection, still remains as a research problem.
    
    \subsection{Applications}
    Automatic Speaker Recognition techniques are transferable to the aforementioned tasks: Language Recognition \citep{dehak2011language}, Age Estimation \citep{chen2018measuring} \citep{Ghahremani2018}, Emotion Classification \citep{Cho2018}, and Spoofing Attacks Detection \citep{lai2018attentive}.

\section{Adapted Gaussian Mixture Models (GMM-UBM)}

In the 1990s, Gaussian Mixture Models (GMM) based systems was the dominant approach to automatic speaker verification. Building on top of GMM, Gaussian Mixture Model-Universal Background Model (GMM-UBM) builds a large speaker-independent GMM, referred to as UBM, and adapts the UBM to specific speaker models via Bayesian adaptation \citep{reynolds2000speaker}. UBM-GMM is the basis for later work such as the i-vectors, which collects sufficient statistics from a UBM, and UBM-GMM is one of the most important developments for automatic speaker verification.

    \subsection{Likelihood Ratio Detector}
    The task of speaker verification is to determine whether an test utterance $U$ is spoken by a given speaker $S$. GMM-UBM defines two models: Background Model (UBM) and Speaker Model (GMM). If the likelihood that $U$ comes from $S$-dependent GMM is larger than the likelihood that $U$ comes from $S$-independent UBM, then $U$ is spoken by $S$, and vice versa. The process above is defined as likelihood ratio:
    \begin{flalign}
    \delta = \frac{P(U\mid GMM)}{P(U\mid UBM)},
    \end{flalign}
    where $\delta$ is called the likelihood ratio detector. Figure \ref{fig:likelihood_ratio_detector} is an illustration of $\delta$.
    
    \begin{knitrout}\definecolor{shadecolor}{rgb}{0.969, 0.969, 0.969}\color{fgcolor}\begin{figure}
    \centering\includegraphics[width=\maxwidth]{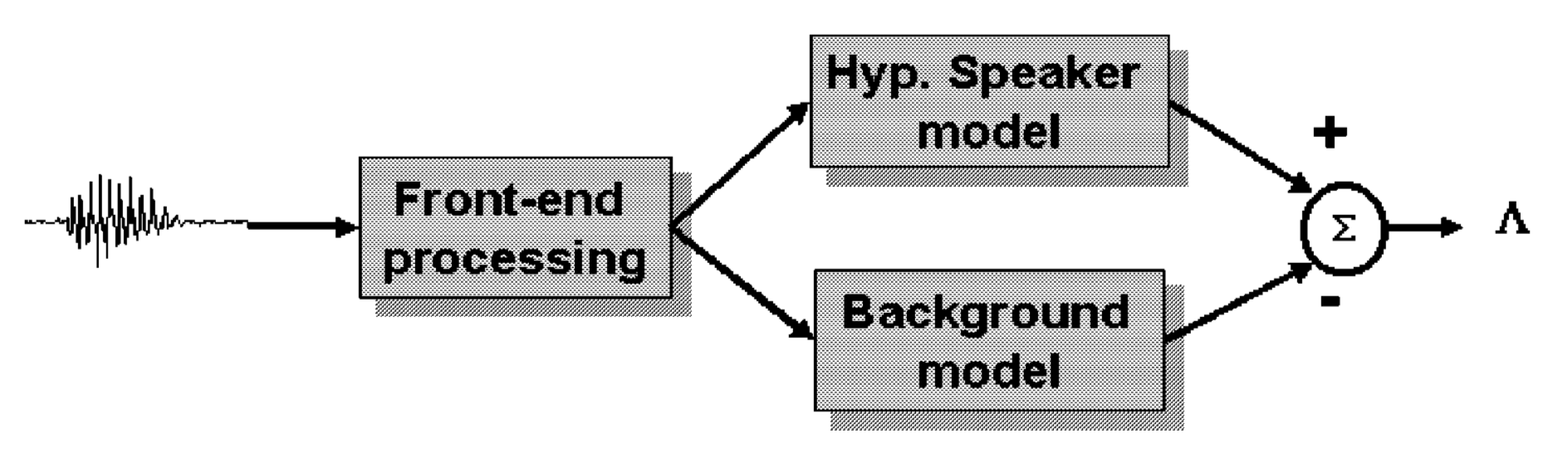} \caption[{\bf Liklihood Ratio Detector for GMM-UBM} ]{{\bf Liklihood Ratio Detector for GMM-UBM \citep{reynolds2000speaker}.}  }\label{fig:likelihood_ratio_detector}
    \end{figure}\end{knitrout}
    
    \subsection{UBM}
    One basic assumption GMM-UBM assumes is that human speech can be decomposed into speaker-independent and speaker-dependent characteristics. Speaker-independent characteristics are traits that are shared across human speech, and example of which could be pitch and vowels. Speaker-dependent characteristics are traits that are unique to every speaker, and example of which could be accent. GMM-UBM builds upon this assumption. First, speaker-independent characteristics are modeled by a large GMM, a UBM. Since it should capture traits shared across all humans, UBM is trained on large data, usually the whole train dataset. Secondly, speaker-dependent characteristics, which is usually presented in the enrollment data, is obtained by adapting the UBM. UBM is trained by the EM algorithm, and the speaker model adaptation is done via MAP estimation. 
    
    Another motivation to split speaker modeling into two steps is that there is often very little enrollment data. For example, setting up smartphones with finger printer readers usually only takes a couple seconds. The enrollment data that is collected is too little to build a powerful model. On the other hand, there are tons of unlabelled data available for training but it does not come from the user. GMM-UBM is one solution that takes advantage of large unlabelled data to build a speaker-specific model by adaptation. 
    
    \subsection{MAP Estimation}
    MAP estimation is illustrated in Figure \ref{fig:adaptation}. Given the sufficient statistics of UBM (mixture weights $w$, mixture means $m$, mixture variances $v$) and some enrollment data, MAP estimation linearly adapts $w$, $m$ and $v$. In \citep{reynolds2000speaker}, all $w$, $m$ and $v$ are adapted although it is common to only adapt the mixture means, and keep the weights and variances fixed.

    \begin{knitrout}\definecolor{shadecolor}{rgb}{0.969, 0.969, 0.969}\color{fgcolor}\begin{figure}
    \centering\includegraphics[width=\maxwidth]{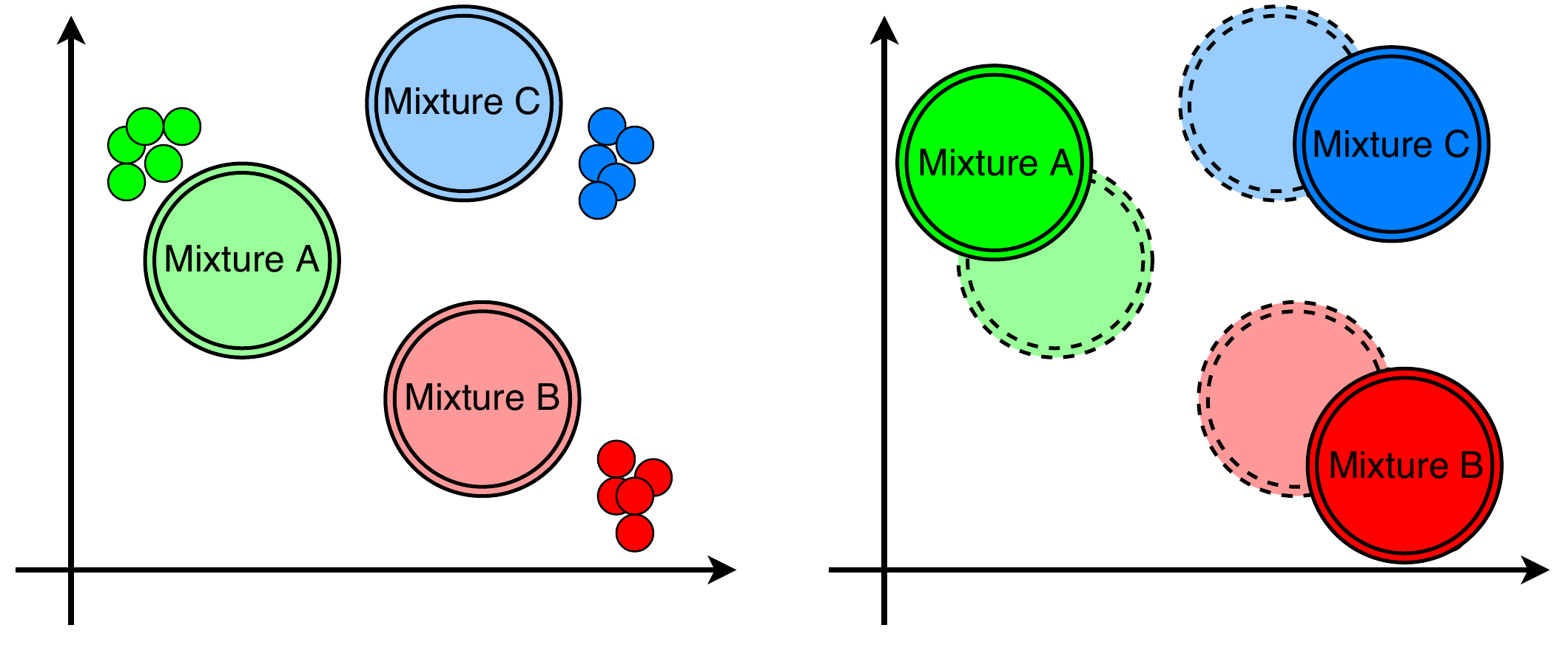} \caption[{\bf GMM-UBM: Adaptation for a Speaker Model} ]{{\bf Speaker Adaptation illustration of GMM-UBM with threee mixtures.} (Left) Universal Background Model with three mixtures and some training data. (Right) Adaptation of speaker model with maximum a posteriori estimation using enrollment data. Note that in this case, only the mixture means are adapted, and mixture variance is fixed.}\label{fig:adaptation}
    \end{figure}\end{knitrout}

\section{Joint Factor Analysis (JFA)}
Joint Factor Analysis is proposed to compensate the shortcomings of GMM-UBM. Refer to Figure \ref{fig:adaptation}, UBM is adapted via MAP to speaker-dependent GMM. If we consider only mean adaptation, we can put the mean vectors $m$ of each Gaussian mixture into a huge vector, which is termed the "Supervector". Let $m\in\mathbb{R}^{F}$, where $F$ is the feature dimension, and assume there are $C$ number of mixtures in the UBM. Then, the supervector $\mathbf{m}\in\mathbb{R}^{F\times C}$. Let us further denote the real speaker mean supervector as $\mathbf{M}$, then MAP estimation is essentially a high-dimensional mapping from $\mathbf{m}$ to $\mathbf{M}$. This is not ideal since MAP not only adapts speaker-specific information but also the channel effects (Figure \ref{fig:speaker_detection_overview}). Another disadvantage of representing speaker with a mean supervector is that the dimension is too huge. For example, it is common to have $F$ as 39 (with delta and double-deltas), and $C$ as 1024. $F\times C$ will end up with a almost 40,000 dimension supervector. 

JFA proposed to address the problem by splitting the supervector $\mathbf{M}$ into speaker independent, speaker dependent, channel dependent, and residual subspaces \citep{lei2011joint}, with each subspace represented by a low-dimensional vector. JFA is formulated as follows: 
\begin{flalign}
\mathbf{M} = \mathbf{m} + Vy + Ux + Dz, 
\end{flalign}\label{jfa_eq}
where $V, U, D$ are low rank matrices for speaker-dependent, channel-dependent, and residual subspaces respectively. With JFA, a low dimensional speaker vector $y$ is extracted. Compare $y$ to GMM-UBM's $M$, $y$ is of much lower dimension (300 v.s. 40,000) and does not have channel effects. 

\section{Front-End Factor Analysis (i-vectors)}
One empirical finding suggested that the channel vector $x$ in JFA also contains speaker information, and a subsequently modification of JFA is proposed and has been one of the most dominant speech representaiton in the last decade: the i-vectors \citep{dehak2011front}. The modified formula is:
\begin{flalign}
\mathbf{M} = \mathbf{m} + Tw, 
\end{flalign}
where $T$ is the total variability matrix (also low rank), and $\textit{w}$ is the i-vectors. Compare this to Equation \ref{jfa_eq}, there is only one low-rank matrix which models both speaker and channel variabilities. Figure \ref{fig:supervectors} is a simple illustration of how JFA and i-vectors converts the supervectors to a low-dimensional embedding. 
\begin{knitrout}\definecolor{shadecolor}{rgb}{0.969, 0.969, 0.969}\color{fgcolor}\begin{figure}
\centering\includegraphics[width=\maxwidth]{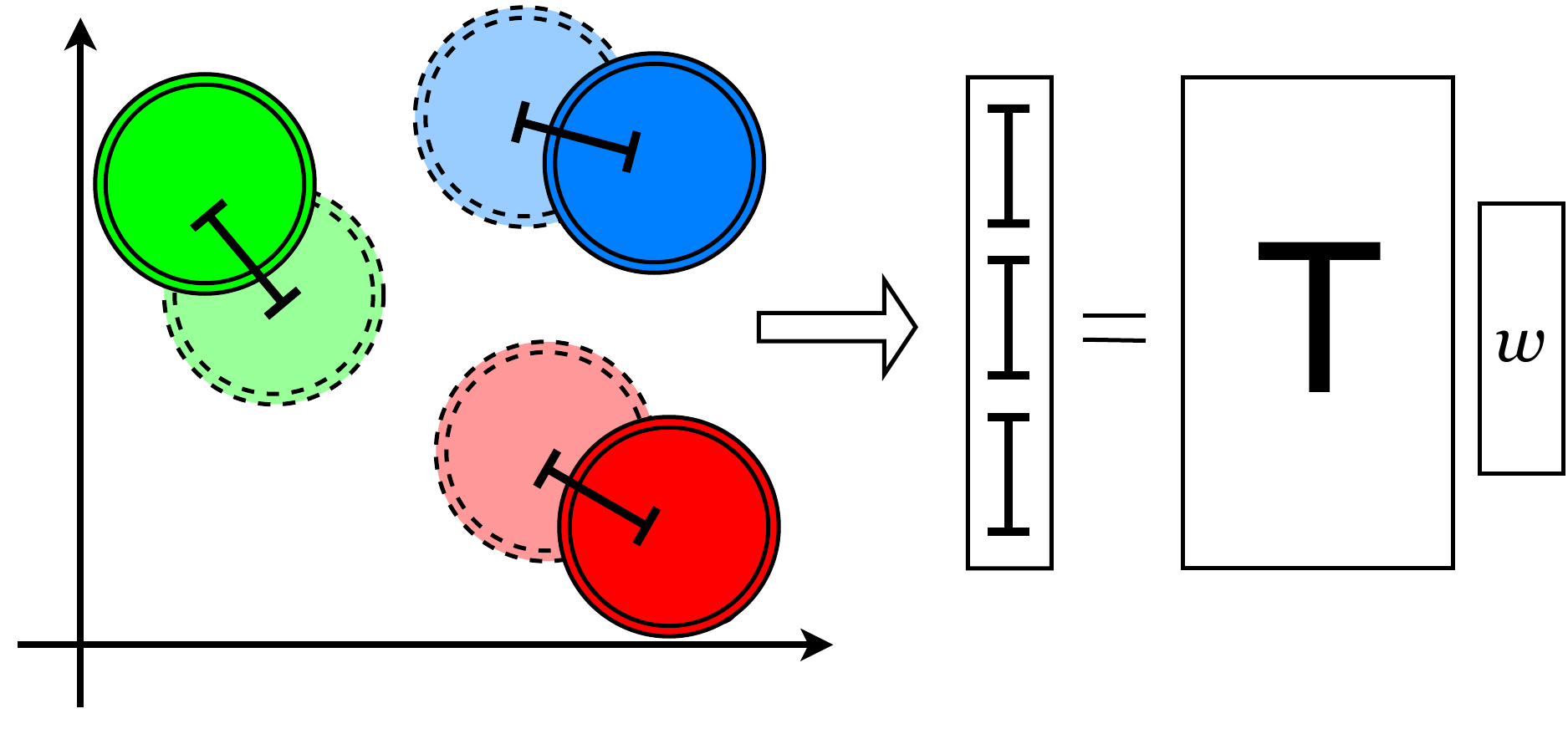} \caption[{\bf GMM-UBM: Adaptation for a Speaker Model} ]{{\bf Speaker Adaptation illustration of GMM-UBM with threee mixtures.} Supervectors}\label{fig:supervectors}
\end{figure}\end{knitrout}    

After $w$ is extracted, it is used to represent the speaker. In Figure \ref{fig:sr_pipeline}, we refer to i-vectors as a summarization step since it reduces the variable-length supervector to a fixed-length vector. In \citep{dehak2011front}, SVM and cosine similarity are used for backend processing. However, i-vector PLDA was a more popular combination. 

\section{Robust DNN Embeddings (x-vectors)}
i-vectors systems have produced several state-of-the-art results on speaker-related tasks. However, as with any statistical systems, an i-vector system is composed of several independent (unsupervised) subsystems trained with different objectives: an UBM for collecting sufficient statistics, an i-vector extrator for extracting i-vectors, and a scoring backend (usually PLDA). x-vectors systems is a supervised DNN-based speaker recognition system that was aimed to combine the clustering and summarization steps in Figure \ref{fig:sr_pipeline} into one
\citep{snyder2017deep}\citep{snyder2018x}. The DNN is based on Network-In-Network \citep{lin2013network}, and trained to classify different speakers (Figure \ref{fig:xvectors}). The layer outputs after the statistical pooling layer can be used as the speaker embeddings, or the x-vectors. Since x-vectors is based on DNN, which requires lots of data, x-vectors systems also utilize data-augmentation by adding noises and reverberations to increase the total amount of data. x-vectors do not necessarily outperform i-vectors on speaker recognition, especially if data and computational resources are limited. 

\begin{knitrout}\definecolor{shadecolor}{rgb}{0.969, 0.969, 0.969}\color{fgcolor}\begin{figure}
\centering\includegraphics[width=0.75\maxwidth]{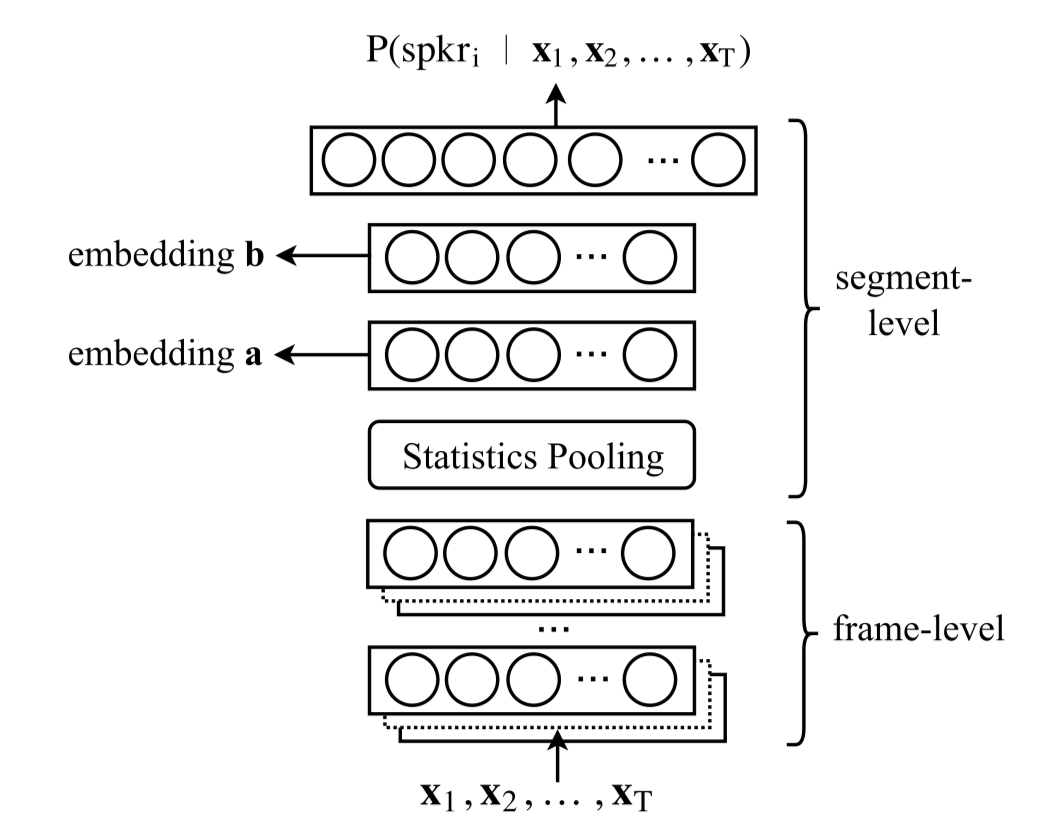} \caption[{\bf x-vectors} ]{{\bf x-vectors \citep{snyder2017deep}.}
}\label{fig:xvectors}
\end{figure}\end{knitrout}

\section{Learnable Dictionary Encoding (LDE)}
The x-vectors framework is not truly end-to-end since it uses a separately trained PLDA for scoring. An elegant end-to-end framework, Learnable Dictionary Encoding, explores a few pooling layers and loss functions \citep{cai2018exploring}, and showed that it is possible to combine the clustering, summarization, and backend processing steps in Figure \ref{fig:sr_pipeline}.

Instead of using a feed-forward deep neural network, LDE employs ResNet34 \citep{he2016deep} in its framework. In addition, contrary to the x-vectors DNN in Figure \ref{fig:xvectors} where there are few layers after the pooling layer, LDE only has a fully-connected layer (for classification) after its pooling layer. LDE uses a LDE layer for pooling (or summarization) in Figure \ref{fig:lde_layer}.   
\begin{knitrout}\definecolor{shadecolor}{rgb}{0.969, 0.969, 0.969}\color{fgcolor}\begin{figure}
\centering\includegraphics[width=0.75\maxwidth]{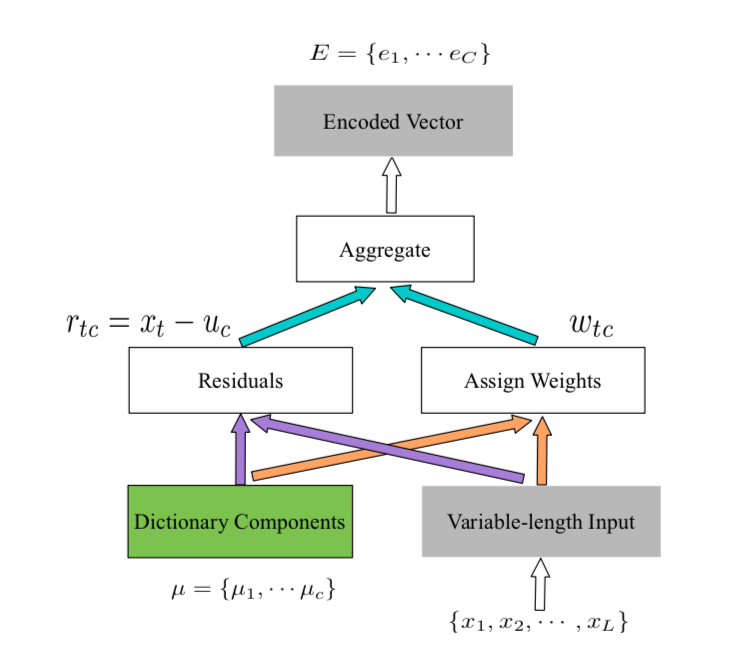} \caption[{\bf Illustration of Learnable Dictionary Encoding layer} ]{{\bf Learnable Dictionary Encoding layer \citep{cai2018exploring}.} LDE layer is inspired from the dictionary-learning procedure of GMM, where a set of dictionary means and weights are learned and aggregated for calculating the fixed-dimensional representation (speaker representation).
}\label{fig:lde_layer}
\end{figure}\end{knitrout}
    
i-vectors and x-vectors systems requires a separately trained backend (PLDA) for scoring, and LDE showed that with Angular Softmax Losses \citep{liu2017sphereface}, a separate backend is not necessary and hence the wholeframework is end-to-end.


\cleardoublepage
\chapter{Conventional Speech Features}
\label{chap:conv_feat}

\section{Introduction}
The Feature Processing step in \ref{fig:sr_pipeline} extracts low-level feature descriptors from raw waveform, and several earlier work showed that Fourier analysis based transforms can effectively capture information of speech signals. Conventional low-level speech features include Log-spectrogram, Log-Filterbank, Mel-Frequency Cepstral Coefficients (MFCC), and Peceptual Linera Predictive (PLP) Analysis. DNN-based speech recognition systems \citep{hinton2012deep}, GMM-UBM systems \citep{reynolds2000speaker} and i-vectors systems \citep{dehak2011front} are based on MFCC; x-vectors systems \citep{snyder2018x} and LDE \citep{cai2018exploring} are based on Log-Filterbank; Attentive Filtering Network \citep{lai2018attentive} is based on Log-Spectrogram. We established our baseline on MFCC, and this chapter will introduce MFCC and the MFCC configuration used in our experiments in Chapter \ref{chap:experiments_results}.

\section{Mel-Frequency Cepstral Coefficients (MFCC)}
MFCC is one of the most standard and common low-level feature in automatic speaker recognition systems. The procedure of MFCC extraction is followed:
\begin{enumerate}
  \item Take Short-Term Fourier Transform (STFT) on the waveform. This step will give us a Spectrogram. 
  \item Apply Mel-scale Filters. This step will give us a Filterbank. 
  \item Take the logarithm on the powers in all Mel-bins. Logarithm is taken also for Log-Spectrogram and Log-Filterbank. 
  \item Apply Discrete Consine Transform (DCT), and keep several cepstral coefficients. This step decorrelates and reduces the dimensionality. 
\end{enumerate}
A visual comparison of Log-Spectrogram, Log-Filterbank, and MFCC is \ref{fig:logspectrogram_filterbank_mfcc}. We can see that there are more structures in Log-Spectrogram and Log-Filterbank, and MFCC has less dimensions than the former two.

\begin{knitrout}
\definecolor{shadecolor}{rgb}{0.969, 0.969, 0.969}\color{fgcolor}\begin{figure}
\centering\includegraphics[width=0.8\maxwidth]{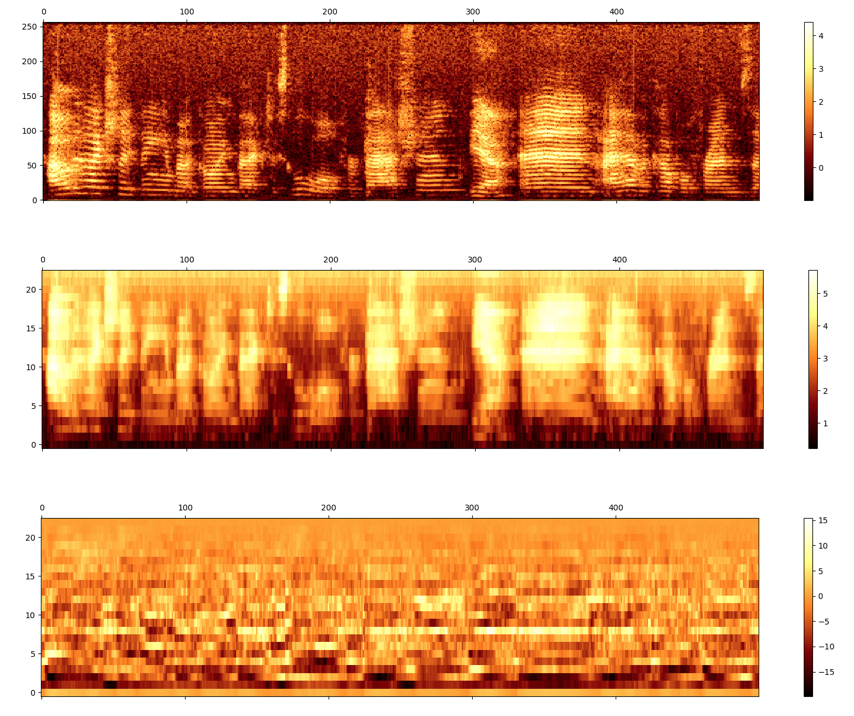} \caption[{\bf Visual Comparison of Log-Spectrogram, Log-Filterbank, and MFCC.} ]{{\bf An Visual Comparison of (top) Log-Spectrogram, (middle) Log-Filterbank, and (bottom) MFCC.}
}\label{fig:logspectrogram_filterbank_mfcc}
\end{figure}\end{knitrout}

\section{MFCC Details}
Our experiments (see Chapter \ref{chap:experiments_results} for more details) are conducted on the LibriSpeech Corpus \citep{panayotov2015librispeech}, in which speech utterances are recorded in 16k Hz. We used the standard 25 ms frame-length and 10 ms frame-shift for STFT computation, 40 Mel filters, and took 24 cepstral coefficients after DCT. The first and second order derivatives (deltas and double-deltas) are computed during UBM training. Details of our MFCC configuration is in Table \ref{table:mfcc_config}.

\begin{table}[]
\centering
\begin{tabular}{|c|c|}
\hline
\multicolumn{2}{|c|}{MFCC Details} \\ \hline
Sampling Frequency & 16000 Hz \\ \hline
Frame Length for STFT & 25 ms \\ \hline
Frame Shift for STFT & 10 ms \\ \hline
High Frequency Cutoff  for Mel Bins & 7600 Hz \\ \hline
Low Frequency Cutoff for Mel Bins & 20 Hz \\ \hline
Number of Mel Bins & 40 \\ \hline
Number of Cepstral Coefficients after DCT & 24 \\ \hline
\end{tabular}
\caption{\bf{Our MFCC Configuration.} The configuration is mostly based on the Kaldi toolkit \citep{povey2011kaldi}.}\label{table:mfcc_config}
\end{table}

\cleardoublepage
\chapter{Contrastive Predictive Coding}
\label{chap:cpc}

\newtheorem{thm}{Theorem} 


\section{Introduction} 
Predictive coding is a well-motivated and developed research area in neuroscience. The central idea of predictive coding is that the current and past states of a system contain relevant information of its future states. On the other hand, one long-standing research question in speech processing has been to extract global information from noisy speech recordings. In speech recognition, this can be related to as retrieving phone labels from the recordings. In speaker recognition, the same research question could be framed as sentiment analysis of the recordings. Could we harness the concept of predictive coding to design a model which extracts representations that are invariant to noise? Contrastive Predictive Coding (CPC) connects the idea of predictive coding and representation learning. This Chapter will give a background overview of predictive coding in neuroscience \ref{sec:sec_pc_neuro}, a background of CPC \ref{sec:sec_cpc_theory} and CPC models \ref{sec:sec_cpc_theory}. Lastly, the application of CPC for speaker verification is presented \ref{sec:sec_cpc_ivector}. 


\section{Predictive Coding in Neuroscience}\label{sec:sec_pc_neuro}
In a famous study by \citep{hubel1968receptive}, the visual Receptive Field (RF) in the monkey striate cortex is studied. Macaque monkey is presented with line stimuli of different orientations while RF responses in the striate cortex are recorded. The experiment showed that cells responded optimally (with high firing rates) to particular line orientations, illustrated in Figure \ref{fig:monkey_striate_cortex}. The interesting question to ask here is: why don't neurons always respond in proportion to the stimulus magnitude? 

\begin{knitrout}
\definecolor{shadecolor}{rgb}{0.969, 0.969, 0.969}\color{fgcolor}\begin{figure}
\centering\includegraphics[width=0.75\maxwidth]{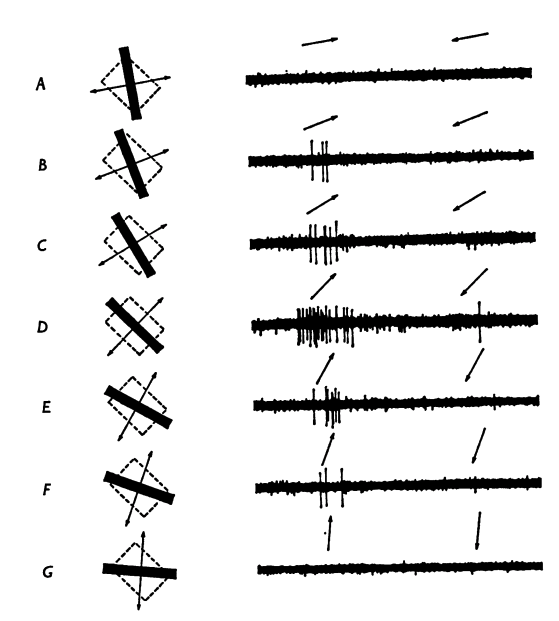} \caption[{\bf Receptive field responses to line stimuli in the monkey striate cortex} ]{{\bf RF responses to line stimuli} Illustration of the RF firing responses to the same line segment but different line orientations from a cell in the monkey striate cortex \citep{hubel1968receptive}}\label{fig:monkey_striate_cortex}
\end{figure}
\end{knitrout}

Predictive coding is one prominent theory that aims to provide a possible explanation. Predictive coding states that human brain can be modeled by a framework that is constantly generating hypotheses and fixing its internal states through an error feedback loop. Since neighboring neurons are likely to be correlated, predictive coding implies that the RF response of a neuron can be predicted by those RF responses of its surroundings, and therefore a strong stimulus does not always correspond to a strong RF response. The first hierarchical model with several levels of predictive coding is proposed for visual processing in \citep{rao1999predictive}. Each level receives a prediction from the previous level and calculates the residual error between prediction and the reality. To achieve efficient coding, only the residual error is propagated forward to the next level, while the next prediction for the current level is made, illustrated in Figure \ref{fig:predictive_coding_visual}. 

\begin{knitrout}
\definecolor{shadecolor}{rgb}{0.969, 0.969, 0.969}\color{fgcolor}\begin{figure}
\centering\includegraphics[width=0.7\maxwidth]{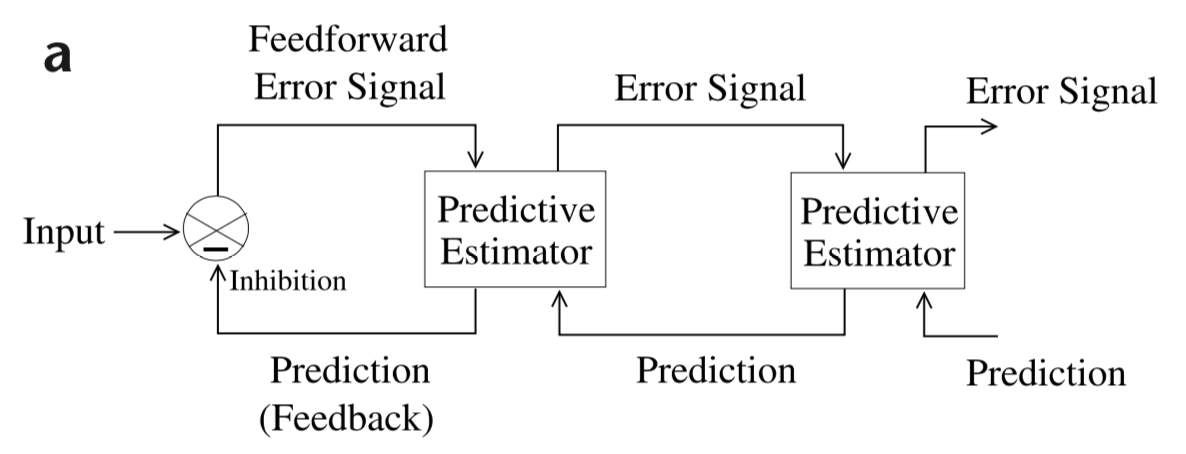} \caption[{\bf Hierarchical model of predictive coding} ]{{\bf Hierarchical model of predictive coding} Illustration of how residual error is propagated and how prediction is made in the hierarchical model in \citep{rao1999predictive} }\label{fig:predictive_coding_visual}
\end{figure}
\end{knitrout}

The study of \citep{rao1999predictive} suggested the importance of feedback connection in addition to feedforward information transmission for visual processing. However, the key insight of how predictive coding is connected to representation learning is that by learning to predict, the model should implicitly retain properties or structures of the input. 



    


\section{Contrastive Predictive Coding (CPC)}\label{sec:sec_cpc_theory}
    
    \subsection{Connection to Predictive Coding}
    Contrastive Predictive Coding (CPC) is proposed in \citep{oord2018representation} as a new unsupervised representation learning framework. One challenging aspect of representation learning within high dimensional signal is noise. The primary goal of CPC is to extract high-level representation, or the slow-varying features \citep{wiskott2002slow}, from a sensory signal full of low-level noises. On the other hand, predictive coding retains properties or structures of the input \ref{sec:sec_pc_neuro}. By predicting the future, the model has to infer global properties or structures from the past, and therefore has to separate global information from noises. One example is TV show series. After watching several episodes of a TV series, most people could generally predict some plots in the next few episodes. But only a few who know the entire series and its history very well can make plot predictions beyond five episodes. These few people has "mastered" the TV series such that they can tell the important plot development from those that are minor in comparison. CPC leverage this idea and therefore could be powerful for separating high-level representation from noises. 
    
    However, how do we quantify high-level representation and monitor how well the model is learning? To quantify high-level representation, CPC calculates the mutual information $I(x;C)$ between the sensory signal $x$ and global information $C$. Let us refer back to the TV series example. The correct prediction of the plots in future episodes are often hidden as several key points in previous episodes. If we put it in terms of mutual information, the sensory signal $x$ is the future episode plots, and global information $C$ is the several key points, such as an important plot twist or character development. \ref{subsec:mi} gives a background of mutual information theory. 
    
    What metric should we use to train the predictive coding model? Figure \ref{fig:predictive_coding_visual} is the original hierarchical model of predictive coding proposed for visual processing, and from the figure we can see that the residual error is calculated during the feedforward pass. An straightforward implementation of residual error could be the L1 loss \ref{l1_loss} or Mean Squared Error (MSE) \ref{mse_loss} between prediction $D(H)$ and actual value $A$, where $H$ is some learnable latent representation and $D$ is a mapping from the latent space to input space. In fact, this implementation can be dated back to the 1960s where MSE is used for training the predictive coding model for speech coding \citep{atal1970adaptive}. Predictive Coding Network, another predictive coding based unsupervised learning framework, is trained with L1 loss \citep{lotter2016deep}. However, either L1 loss or MSE loss requires a mapping function, namely a decoder $D$, that computes $p(x\mid C)$. In our TV series example, $p(x\mid C)$ is saying, "tell me all the details $x$ of future plots given the several key points $C$. Intuitively, this is a hard task and unnecessary for our purpose since we are interested in high-level representations. To get around this issue, CPC models the mutual information directly with the noise contrastive estimation technique, which is introduced in \ref{subsec:nce}.

    \begin{flalign}
    L1 = \sum_{i=1}^N (D(h_{i})-a_{i})^2
    \end{flalign}\label{l1_loss}.
    \begin{flalign}
    MSE = \sum_{i=1}^N \mid D(h_{i})-a_{i}\mid.
    \end{flalign}\label{mse_loss}

    \begin{knitrout}
    \definecolor{shadecolor}{rgb}{0.969, 0.969, 0.969}\color{fgcolor}\begin{figure}
    \centering\includegraphics[width=0.9\maxwidth]{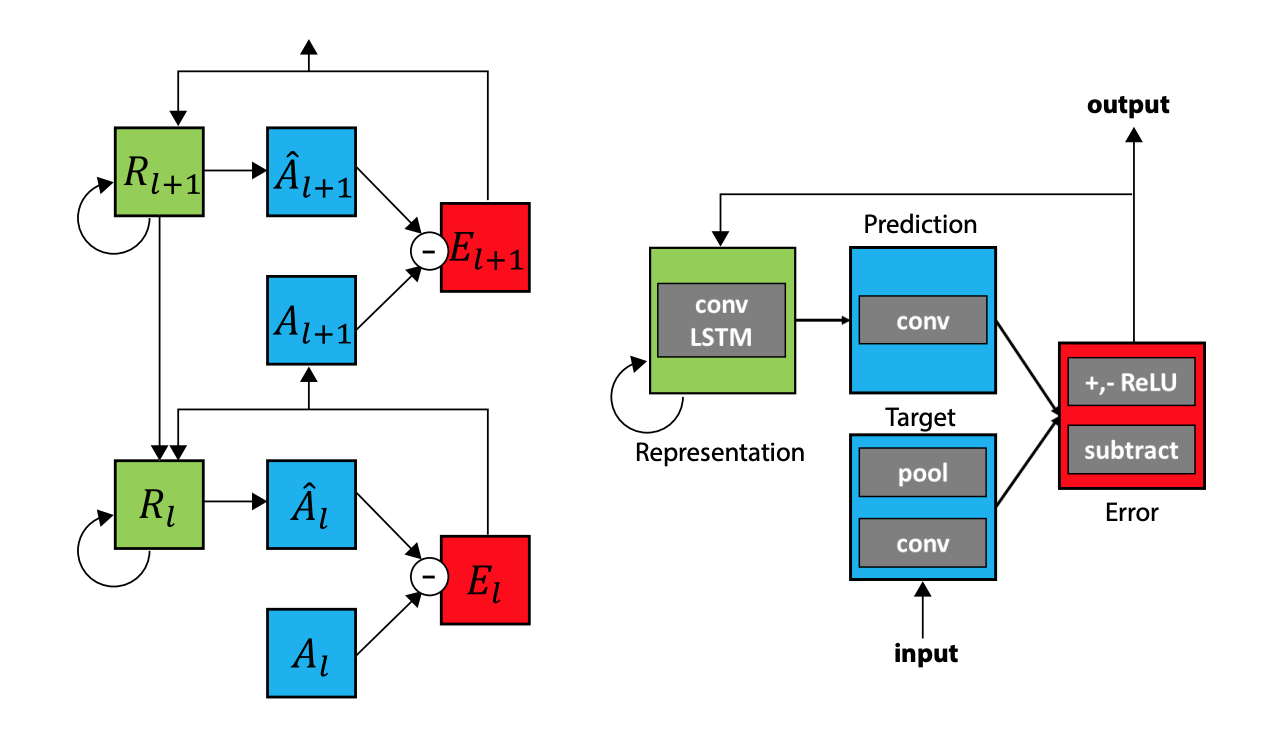} \caption[{\bf Predictive Coding Network} ]{{\bf Predictive Coding Network (PredNet)} Illustration of information flow in PredNet, which is trained with the L1 loss between $\hat{A_{l+1}}$ and $A_{l+1}$. \citep{lotter2016deep} }\label{fig:prednet}
    \end{figure}
    \end{knitrout}
    
    \subsection{Mutual Information}\label{subsec:mi}
    Mutual information denotes the amount of information shared between the two variables. Given two random variable $X$ and $Y$, mutual information $I(X;Y)$ is defined as, 
    \begin{flalign}
    I(X;Y) = H(X) - H(X\mid Y), 
    \end{flalign}\label{mutual_information}
    where $H(X)$ is the entropy of $X$ and $H(X\mid Y)$ is the conditional entropy of $Y$ given $X$. $H(X)$ is defined as,
    \begin{flalign}
    H(X) = -\sum_{i=1}^n P(X=x_{i})\log{P(X=x_{i})}, 
    \end{flalign}\label{eq:entropy} 
    and $H(X\mid Y)$ is defined as, 
    \begin{flalign}
    H(X\mid Y) = -\sum_{i=1}^n P(X=x_{i}\mid Y)\log{P(X=x_{i}\mid Y)}.
    \end{flalign}\label{eq:conditional_entropy}    
    With the above definitions, we can subsequently show the following:
    \begin{flalign}
    I(X;Y) = \sum_{i=1}^n\sum_{j=1}^m p(x_{i},y_{j})\log{\frac{p(x_{i}\mid y_{j})}{p(x_{i})}}
    \end{flalign}
    \begin{proof}
    First we expand \ref{eq:conditional_entropy} as: 
    \begin{flalign}
    H(X\mid Y) &= -\sum_{i=1}^n P(X=x_{i}\mid Y)\log{P(X=x_{i}\mid Y)} \\
               &= -\sum_{i=1}^n\sum_{j=1}^m P(X=x_{i}\mid Y=y_{j})P(Y=y_{j})\log{P(X=x_{i}\mid Y=y_{j})} \\
               &= -\sum_{i=1}^n\sum_{j=1}^m p(x_{i}\mid y_{j})p(y_{j})\log{p(x_{i}\mid y_{j})}
    \end{flalign}
    Then by substitution and Baye's rule, 
    \begin{flalign}
    I(X;Y) &= H(X) - H(X\mid Y) \\
           &= -\sum_{i=1}^n p(x_{i})\log{p(x_{i})} + \sum_{i=1}^n\sum_{j=1}^m p(x_{i}\mid y_{j})p(y_{j})\log{p(x_{i}\mid y_{j})} \\
           &= -\sum_{i=1}^n\sum_{j=1}^m p(x_{i},y_{j})\log{p(x_{i})} + \sum_{i=1}^n\sum_{j=1}^m p(x_{i},y_{j})\log{\frac{p(x_{i},y_{j})}{p(y_{j})}}\\
           &= -\sum_{i=1}^n\sum_{j=1}^m p(x_{i},y_{j})\log{\frac{p(x_{i})p(y_{j})}{p(x_{i},y_{j})}}\\
           &= \sum_{i=1}^n\sum_{j=1}^m p(x_{i},y_{j})\log{\frac{p(x_{i}\mid y_{j})}{p(x_{i})}}
    \end{flalign}
    \end{proof}
    We can also easily show that if $X$ and $Y$ are independent, their mutual information is zero: 
    \begin{proof}
    Given $X$ and $Y$ are independent, $P(X\mid Y) = P(X)$. By definition, we can rewrite $H(X\mid Y)$ as:  
    \begin{flalign}
    H(X\mid Y) &= -\sum_{i=1}^n P(X=x_{i}\mid Y)\log{P(X=x_{i}\mid Y)} \\
               &= -\sum_{i=1}^n P(X=x_{i})\log{P(X=x_{i})} \\
               &= H(X),               
    \end{flalign}
    and therefore, we have:
    \begin{flalign}
    I(X;Y) &= H(X) - H(X\mid Y) \\
           &= H(X) - H(X) \\
           &= 0 
    \end{flalign}
    \end{proof}
    
    In the context of representation learning, mutual information gives us a quantitative measure of how well a model learns the global information. Let us look back at the TV series example again. If a person only has limited memory and has successfully observed the key developments, denoted as $C_{1}$, over the past episodes, those developments are likely to be highly relevant to the upcoming episodes, denoted as $X$. We can say that their mutual information $I(X;C_{1})$ is high. Hoewver, given the limited amount of memory everyone has, if the person only remembered the minor plot developments, denoted as $C_{2}$, the mutual information $I(X;C_{2})$ is most likely to be low. 
    
    \subsection{Noise-Contrastive Estimation (NCE)}\label{subsec:nce}
    Noise-Contrastive Estimation (NCE) is an estimation technique for estimating the parameters of parametric density functions \citep{gutmann2012noise}. Let us consider a set of observations $X = (\vec{x_{1}}, \vec{x_{2}}, \vec{x_{3}},..., \vec{x_{N}})$, where $\vec{x_{i}} \in {\mathbb{R}}^n$. In real world examples, $n$ is often of high dimension, and the goal of all machine learning models is to find, or give an accurate estimate of, the underlying data distribution, the probability density function (pdf) $P_{D}$, from the observable set $X$. NCE makes an assumption that $P_{D}$ comes from a parameterized family of functions: 
    \begin{flalign}
    P_{D} \in \{P_{M}(;\pmb{\theta})\},
    \end{flalign}
    where $\pmb{\theta}$ is a set of parameters. Put it another way, there exists some $\theta^{\star}$ such that the following is true, 
    \begin{flalign}
    P_{D} = P_{M}(;\theta^{\star}).
    \end{flalign}
    Now, let us denote any estimate of $\theta^{\star}$ as $\bar{\theta}$. Then, the following must hold for any pdf $P_{M}(;\bar{\theta})$:
    \begin{flalign}
    P_{M}(;\bar{\theta}) \geq 0
    \end{flalign}   
    \begin{flalign}
    \int P_{M}(\vec{x};\bar{\theta}) d\vec{x} = 1
    \end{flalign}  
    If these two constraints are satisfied for all $\theta\in\pmb{\theta}$, then we say $P_{D}$ is normalized; otherwise, $P_{D}$ is unnormalized. It is common for models to be unnormazlied, such as the Gibbs distribution. Let us further give these unnormalized parametric models a name, $P^{0}_{M}(;\alpha)$. To normalize $P^{0}_{M}(;\alpha)$, we would need to calculate the partition function $Z(\alpha)$:
    \begin{flalign}
    Z(\alpha) = \int P^{0}_{M}(\vec{x};\alpha) d\vec{x}, 
    \end{flalign} 
    and $P^{0}_{M}(;\alpha)$ can be normalized by  $\frac{P^{0}_{M}(;\alpha)}{Z(\alpha)}$.

    Everything so far is reasonable, except that in real word examples, $Z(\alpha)$ is certainly intractable for high-dimensional data (curse of dimensionality), and thus $P^{0}_{M}(;\alpha)$ is still unnormalized. One simple solution NCE proposed is, why not make $Z(\alpha)$ an additional parameter \citep{gutmann2012noise}? Let us define the new pdf $P_{M}(;\vec{\theta})$ accordingly: 
    \begin{flalign}
    \ln P_{M}(;\vec{\theta}) := \ln P^{0}_{M}(;\alpha) + c, 
    \end{flalign}\label{eq:normalizaed_scaling}
    where $c = \frac{1}{Z(\alpha)}$, and $\vec{\theta} = (\alpha, c)$. The estimate $\bar{\theta} = (\bar{\alpha}, \bar{c})$ now is not subject to the two constraints above since $\bar{c}$ provides a scaling factor. The intuition here is that instead of calculating $Z(\alpha)$ to normalize $P^{0}_{M}(;\alpha)$ for all $\alpha$, only $P^{0}_{M}(;\bar{\alpha})$ is normalized. 
    
    However, Maximum Likelihood Estimation only works for normalized pdf, and $P^{0}_{M}(;\alpha)$ is not normalized for all $\alpha$. NCE is therefore proposed for estimating \textbf{unnormalized} parametric pdfs. 
    
    \subsubsection{Density Estimation in a Supervised Setting}
    The goal of density estimation is to give an accurate description of the underlying probablistic density distribution of an observable data set $X$ with unknown density $P_{D}$. The intuition of NCE is that by comparing $X$ against a known set $Y$, which has a known density $P_{N}$, we can get a good grasp of what $P_{D}$ looks like. Put it more concretely, by drawing samples from $Y = (\vec{y_{1}}, \vec{y_{2}}, \vec{y_{3}},..., \vec{y}_{T_{y}})$ with a known pdf $P_{N}$, and samples from $X = (\vec{x_{1}}, \vec{x_{2}}, \vec{x_{3}},..., \vec{x}_{T_{x}})$, we can estimate the density ratio $\frac{P_{D}}{P_{N}}$. With $\frac{P_{D}}{P_{N}}$ and $P_{N}$, we have the target density $P_{D}$. 
    
    By classifying samples $X$ from noise $Y$ with a simple classifier, in this case logistic regression, we show NCE gets a estimate of the probability density ratio $\frac{P_{D}}{P_{N}}$.
    
    Let $X$ and $Y$ be two observable sets containing data $X = (\vec{x_{1}}, \vec{x_{2}}, \vec{x_{3}},..., \vec{x}_{T_{x}})$, $Y = (\vec{y_{1}}, \vec{y_{2}}, \vec{y_{3}},..., \vec{y}_{T_{y}})$, and let $U$ be $X\cup Y$, $U = (\vec{u_{1}}, \vec{u_{2}}, \vec{u_{3}},..., \vec{u}_{T_{x}+T_{y}})$. $X$ is drawn from an unknown pdf $P_{D} \in \{P_{M}(;\pmb{\theta})\}$, and $Y$ is drawn from a known pdf $P_{N}$. Since $Y$ is not our target, it is commonly referred to as the "noise". We also assign each datapoint in $U$ a label $C_{t}$: $C_{t} = 1$ if $u_{t}\in X$ and $C_{t} = 0$ if $u_{t}\in Y$. From the above settings, the likelihood distributions are then:
    \begin{flalign}
    P(\vec{u}\mid C=1) = P_{M}(\vec{u};\pmb{\theta})
    \end{flalign}   
    \begin{flalign}
    P(\vec{u}\mid C=0) = P_{N}(\vec{u})
    \end{flalign}
    The prior distributions are: 
    \begin{flalign}
    P(C=1) = \frac{T_{x}}{T_{x} + T_{y}}
    \end{flalign}   
    \begin{flalign}
    P(C=0) = \frac{T_{y}}{T_{x} + T_{y}}
    \end{flalign} 
    The probability of the data $P(\vec{u})$ is thus: 
    \begin{flalign}
    P(\vec{u}) &= P(C=0)\times P(\vec{u}\mid C=0) + P(C=1)\times P(\vec{u}\mid C=1)\\
               &= \frac{T_{y}}{T_{x} + T_{y}}\times P_{N}(\vec{u}) + \frac{T_{x}}{T_{x} + T_{y}}\times P_{M}(\vec{u};\pmb{\theta})
    \end{flalign} 
    With Baye's rule, we can derive the posterior distributions of $P(C=1\mid\vec{u})$ and  $P(C=0\mid\vec{u})$: 
    \begin{flalign}
    P(C=1\mid\vec{u}) &= \frac{P(C=1)\times P(\vec{u}\mid C=1)}{P(\vec{u})}\\
                    &= \frac{\frac{T_{x}}{T_{x} + T_{y}}\times P_{M}(\vec{u};\pmb{\theta})}{\frac{T_{y}}{T_{x} + T_{y}}\times P_{N}(\vec{u}) + \frac{T_{x}}{T_{x} + T_{y}}\times P_{M}(\vec{u};\pmb{\theta})}\\
                    &= \frac{P_{M}(\vec{u};\pmb{\theta})}{P_{M}(\vec{u};\pmb{\theta})+v P_{N}(\vec{u})}
    \end{flalign}  
    where 
    \begin{flalign}
    v = \frac{T_{y}}{T_{x}}.
    \end{flalign} 
    Similarly, we can get
    \begin{flalign}
    P(C=0\mid\vec{u}) = \frac{v P_{N}(\vec{u})}{P_{M}(\vec{u};\pmb{\theta})+v P_{N}(\vec{u})}
    \end{flalign} 
    $P(C=1\mid\vec{u})$ can further be expressed as, 
    \begin{flalign}
    P(C=1\mid\vec{u}) &= \frac{P_{M}(\vec{u};\pmb{\theta})}{P_{M}(\vec{u};\pmb{\theta})+v P_{N}(\vec{u})} \\
                    &= \Big(1+v\frac{P_{N}(\vec{u})}{P_{M}(\vec{u};\pmb{\theta})}\Big)^{-1}
    \end{flalign} 
    Now, we can denote our target density ratio $\frac{P_{N}(\vec{u})}{P_{M}(\vec{u};\pmb{\theta})}$ with a new variable $G$: 
    \begin{flalign}
    G(\vec{u};\pmb{\theta}) &= \ln\frac{P_{M}(\vec{u};\pmb{\theta})}{P_{N}(\vec{u})}\\
                            &= \ln{P_{M}(\vec{u};\pmb{\theta})} - \ln{P_{N}(\vec{u})}.  
    \end{flalign} 
    $P(C=1\mid\vec{u})$ is then:
    \begin{flalign}
    P(C=1\mid\vec{u}) &= \text{sigmoid}(G(\vec{u};\pmb{\theta}))\\
                    &= h(\vec{u};\pmb{\theta})
    \end{flalign} 
    
    Finally, since $C_{t}$ is a Bernoulli distribution with value of $0$ or $1$. We can write the log-likelihood as:
    \begin{flalign}
    l(\pmb{\theta}) &= \sum_{t=1}^{T_{x}+T_{y}} C_{t}\ln P(C_{t}=1\mid\vec{u_{t}}) + (1-C_{t})\ln P(C_{t}=0\mid\vec{u_{t}})\\
                    &= \sum_{t=1}^{T_{x}} \ln h(\vec{x_{t}};\pmb{\theta}) + \sum_{t=1}^{T_{y}} \ln \Big(1-h(\vec{y_{t}};\pmb{\theta})\Big)
    \end{flalign}\label{eq:log-likelihood}
    Optimize $l(\pmb{\theta})$ with respect to the parameters $\pmb{\theta}$ will lead to an estimate of $G(\vec{u};\bar{\theta})$, which is the density ratio we want. If we take a step back, we can see that $-l(\pmb{\theta})$ is in fact a cross-entropy loss. In a supervised setting, NCE gives us a density estimation! 
    
    \subsubsection{The NCE Estimator}
    Let us refer back to \ref{eq:normalizaed_scaling}. We are now ready to introduce the NCE estimator:
    \begin{flalign}
    J_{T}(\vec{\theta}) = \frac{1}{T_{d}}\Bigg(\sum_{t=1}^{T_{x}} \ln h(\vec{x_{t}};\vec{\theta}) + \sum_{t=1}^{T_{y}} \ln \Big(1-h(\vec{y_{t}};\vec{\theta})\Big)\Bigg),
    \end{flalign}\label{eq:nce_estimator}
    which is off by a scaling constant as \ref{eq:log-likelihood}.
    

\section{Representation Learning with CPC}\label{sec:sec_cpc_implement}
    
    \subsection{Single Autoregressive Model}
    As mentioned in the previous sections, mutual information gives the model a good criterion to measure how much global information is preserved. We can explicitly write out the formula for mutual information:
    \begin{flalign}
    I(X;Y) = \sum_{i=1}^n\sum_{j=1}^m p(x_{i},y_{j})\log{\frac{p(x_{i}\mid y_{j})}{p(x_{i})}}
    \end{flalign}
    In speech, we can make $X$ the waveform of any utterance, and $Y$ the global information such as speaker label. Therefore, the mutual information we are interested in becomes:
    \begin{flalign}
    I(U;S) = \sum_{i=1}^n\sum_{j=1}^m p(u_{i},s_{j})\log{\frac{p(u_{i}\mid s_{j})}{p(u_{i})}}
    \end{flalign}
    where $U$ represents utterance and $S$ represents speaker label. In \citep{oord2018representation}, NCE objective is introduced for model training, and the term $\frac{p(u_{i}\mid s_{j})}{p(u_{i})}$ is selected as the density ratio to be estimated in NCE. We will prove why $\frac{p(u_{i}\mid s_{j})}{p(u_{i})}$ is selected later. The NCE objective is subsequently named NCE loss. 
    
    \ref{fig:basic_cpc_model_train} is an illustration of the proposed CPC model in \citep{oord2018representation}. The model takes in raw waveforms $U$ as input and transforms it to some latent space $L$ by an encoder. In the latent space, an Recurrent Neural Network is trained by the NCE loss to learn $S$.  
    
    \subsubsection{NCE Loss}
    CPC selects $\frac{p(u_{i}\mid s_{i})}{p(u_{i})}$ as the density ratio to be estimated in the NCE estimator. We can denote it with $f_{i}$:
    \begin{flalign}
    f_{i}(u_{i},s_{i}) = \frac{p(u_{i}\mid s_{i})}{p(u_{i})}
    \end{flalign}
    We can see that $f_{i}$ is unnormalized, and this is the reason why we started off with NCE. In addition, since $f_{i}$ could not be explicitly computed. An alternative way is to model $f_{i}$ with log-bilinear model, which signifies how relevant the input is to the context:
    \begin{flalign}
    f_{i}(u_{i},s_{i}) = \exp{(s_{i}\cdot u_{i})}.
    \end{flalign}   
    Refer back to the model \ref{fig:basic_cpc_model_train}, we can see that $s_{i}$ is modeled by the context vector $C_{i}$ of the recurrent neural network, and $u_{i}$ can be modeled by either the waveform or latent space $L_{i}$. Since we would like the model to learn high-level information, it makes more sense to model $u_{i}$ with $L_{i}$. Therefore, $f_{i}$ becomes: 
    \begin{flalign}
    f_{i}(u_{i},s_{i}) = \exp{(C_{i}\cdot L_{i})}. 
    \end{flalign}\label{eq:original_density_ratio}
    However, the dimension of the context vector $C_{i}$ and latent space $L_{i}$ do not always agree. A simple solution is to add a matrix to conform the dimension. Let $C_{i}\in {\mathbb{R}}^{a}$ and $L_{i}\in {\mathbb{R}}^{b}$. We define a matrix $W_{i}\in {\mathbb{R}}^{a\times b}$ and \ref{eq:original_density_ratio} becomes:
    \begin{flalign}
    f_{i}(u_{i},s_{i}) &= \exp{\big(L_{i}\cdot (W_{i}C_{i})\big)}\\
                    &= \exp{\big(L_{i}^{T}(W_{i}C_{i})\big)}
    \end{flalign}\label{eq:modified_density_ratio}
    We are now ready to define the NCE loss $\mathcal{L}$ for training the CPC model. Refer to \ref{eq:nce_estimator}, NCE gives an estimate of the density ratio by classifying data samples from noise samples. Given a batch of utterances $B = (b_{1}, b_{2}, b_{3}, ..., b_{N})$, which includes $1$ data sampels and $N-1$ noise samples, where the positive sample comes from the data distribution $p(u_{i}\mid s_{i})$ and the noise samples come from noise distributions $p(u_{i})$. NCE loss is defined as:
    \begin{flalign}
    \mathcal{L} &= -\frac{1}{N}\sum_{B} \big(\log{\frac{f_{p}(u_{i},s_{i})}{\sum_{B} f_{n}(u_{i},s_{i})}}\big)\\
                &= -\mathop{\mathbb{E}}_{B} \big[\log{\frac{f_{p}(u_{i},s_{i})}{\sum_{B} f_{n}(u_{i},s_{i})}}\big]
    \end{flalign}
    where $u_{i}$ is any frame segment from utterance $b_{i} \forall i$, $s_{i}$ is the corresponding global context for frame segment $u_{i}$, $\frac{f_{p}}{\sum_{B} f_{n}}$ is the prediction of the model, and $\log{\frac{f_{p}}{\sum_{B} f_{n}}}$ is taking the softmax over $B$.  
    
    However, the current loss $\mathcal{L}$ has nothing to do with predictive coding \ref{fig:predictive_coding_visual}, where a prediction of the future is made by the context and the residual error is propagated back to correct the context \citep{lotter2016deep}. Similarly, CPC model also incorporates future frame predictions. We can modify the $\mathcal{L}$ as: 
    \begin{flalign}
    \mathcal{L} = -\mathop{\mathbb{E}}_{B} \mathop{\mathbb{E}}_{T} \big[\log{\frac{f_{p}(u_{i+t},s_{i})}{\sum_{B} f_{n}(u_{i+t},s_{i})}}\big],
    \end{flalign}
    where instead of computing loss only with the density ratio of current frame $f_{i}(u_{i},s_{i})$, we also calculate the density ratio of future frames up to $T$ frames in the future, $f_{i}(u_{i+t},s_{i})$.
    
    \begin{knitrout}
    \definecolor{shadecolor}{rgb}{0.969, 0.969, 0.969}\color{fgcolor}\begin{figure}
    \centering\includegraphics[width=1.\maxwidth]{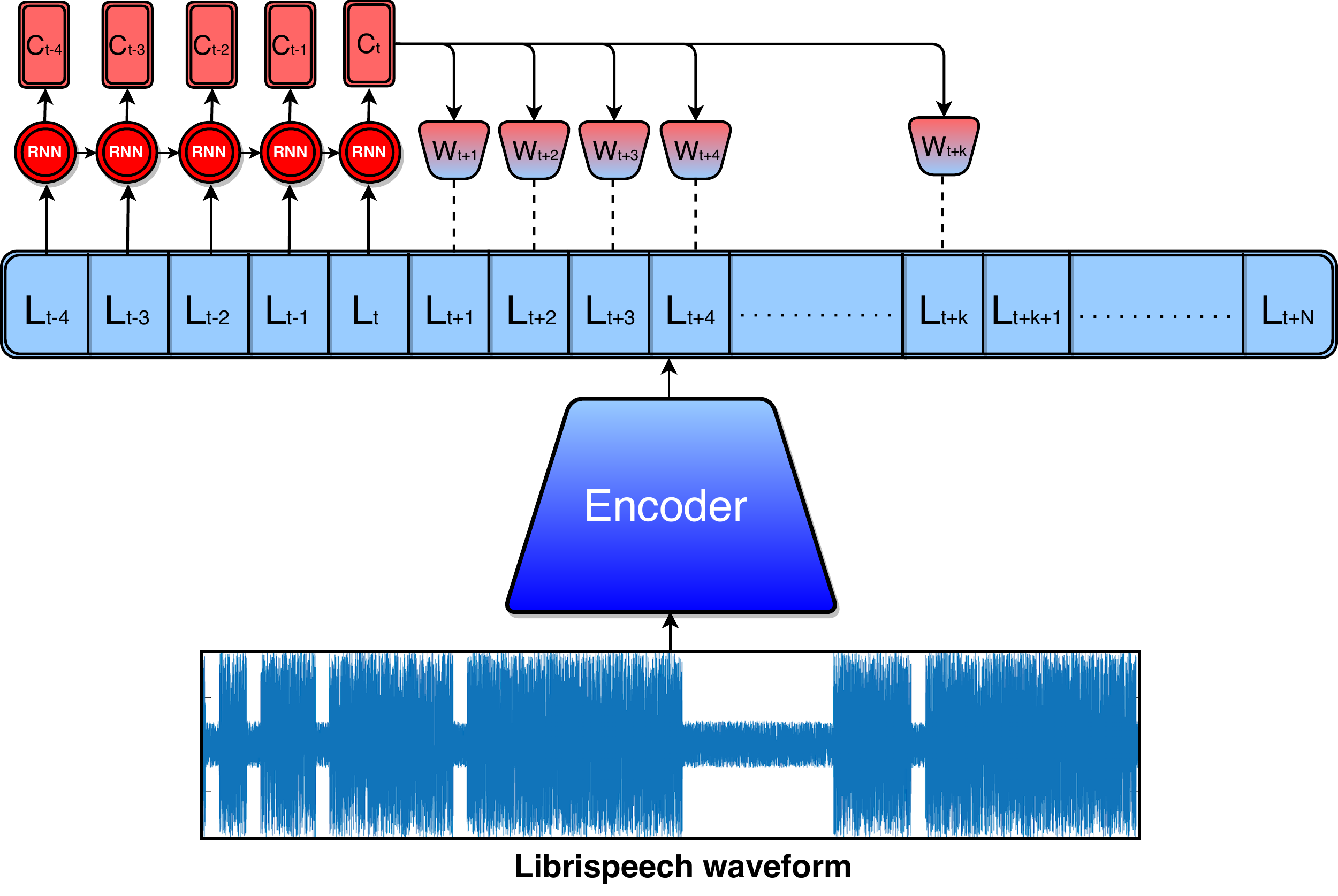} \caption[{\bf CPC Single Autoregressive Model} ]{{\bf CPC Single Autoregressive Model} Illustration of the CPC single autoregressive model's training stage. The model takes in raw waveform and transform it to some latent space by an encoder. An recurrent neural network is trained to learn global information in the latent space with NCE loss. }\label{fig:basic_cpc_model_train}
    \end{figure}
    \end{knitrout}
    
    \subsubsection{Connection to Mutual Information}
    Why does CPC selects $\frac{p(u_{i}\mid s_{i})}{p(u_{i})}$ as the density ratio to be estimated in the NCE estimator? How does it connect to mutual information? 
   
    We will show that minimizing the NCE loss $\mathcal{L}$ will result in maximizing the mutual information. First, we prove that optimizing $\mathcal{L}$ will converge the density ratio $f_{i}(u_{i},s_{i})$ to $\frac{p(u_{i}\mid s_{i})}{p(u_{i})}$. 
    \begin{proof}
    $f_{i}(u_{i},s_{i})$ will converge to $\frac{p(u_{i}\mid s_{i})}{p(u_{i})}$ by optimizing $\mathcal{L}$, where $p(u_{i}\mid s_{i})$ is the data distribution and $p(u_{i})$ is the noise distribution. 
    
    The prediction of $\mathcal{L}$ is $\frac{f_{p}}{\sum_{B} f_{n}}$. Let us denote the optimal probability of classifying positive samples $i$ correctly as $P(i=\text{positive}\mid U,C)$ (it is correct if it comes from the data distribution, and therefore incorrect if it comes from the noise distribution):
    \begin{flalign}
    P(i=\text{positive}\mid U,C) &= \frac{p(u_{i}\mid C)\prod_{j\neq i}{p(u_{j})}}{\sum_{k=1}^{N}{p(u_{k}\mid C)\prod_{j\neq k}{p(u_{j})}}}\\
            &= \frac{\frac{p(u_{i}\mid C)}{p(u_{i})}}{\sum_{k=1}^{N}\frac{p(u_{k}\mid C)}{p(u_{k})}}
    \end{flalign}
    Compare $\frac{f_{p}}{\sum_{B} f_{n}}$ and $P(i=\text{positive}\mid U,C)$ we have,
    \begin{flalign}
    \frac{f_{p}}{\sum_{B} f_{n}} = \frac{\frac{p(u_{i}\mid C)}{p(u_{i})}}{\sum_{k=1}^{N}\frac{p(u_{k}\mid C)}{p(u_{k})}} 
    \end{flalign}
    Therefore, $f_{i}$ will converge to $\frac{p(u_{i}\mid s_{i})}{p(u_{i})}$.
    \end{proof}
    Now, with the optimal $f_{i}$, we can proof mutual information $I(u_{i+t}, s_{i}) >= \log{N} - \mathcal{L}^{opt}$, where $\mathcal{L}^{opt}$ is the optimal loss. Minimizing the NCE loss $\mathcal{L}$ will result in maximizing the mutual information $I(u_{i+t}, s_{i})$. 
    \begin{proof}
    The lower bound for $I(u_{i+t}, s_{i})$ is $\log{N} - \mathcal{L}^{opt}$.
    
    We first rewrite $\mathcal{L}$ by separating the positive sample and negative samples explicitly,   
    \begin{flalign}
    \mathcal{L} &= -\mathop{\mathbb{E}}_{B} \mathop{\mathbb{E}}_{T}\big[\log{\frac{f_{p}(u_{i+t},s_{i})}{\sum_{B}f_{n}(u_{i+t},s_{i})}}\big] \\
                &= -\mathop{\mathbb{E}}_{B} \mathop{\mathbb{E}}_{T}\big[\log{\frac{f_{p}(u_{i+t},s_{i})}{f_{p}(u_{i+t},s_{i}) + \sum_{B_{negative}}f_{n}(u_{i+t},s_{i})}}\big]
    \end{flalign}
    where $B_{negative}$ is the negative samples in batch $B$, in which there are $N$ samples. By substituting the optimal density ratio $f_{i}$ in $\mathcal{L}$, we will get the optimal loss $\mathcal{L}^{opt}$:
    \begin{flalign}
    \mathcal{L}^{opt} &= -\mathop{\mathbb{E}}_{B} \mathop{\mathbb{E}}_{T}\big[\log{\big(\frac{\frac{p(u_{i+t}\mid s_{i})}{p(u_{i+t})}}{\frac{p(u_{i+t}\mid s_{i})}{p(u_{i+t})} + \sum_{B_{negative}}\frac{p(u_{i+t}\mid s_{i})}{p(u_{i+t})}}\big)}\big]\\
                    &= \mathop{\mathbb{E}}_{B} \mathop{\mathbb{E}}_{T}\big[\log{\big(1+\frac{p(u_{i+t})}{p(u_{i+t}\mid s_{i})}\sum_{B_{negative}}\frac{p(u_{i+t}\mid s_{i})}{p(u_{i+t})}\big)}\big] \\
                    &\approx \mathop{\mathbb{E}}_{B} \mathop{\mathbb{E}}_{T}\big[\log{\big(1+\frac{p(u_{i+t})}{p(u_{i+t}\mid s_{i})}(N-1)\mathop{\mathbb{E}}_{B_{negative}}[\frac{p(u_{i+t}\mid s_{i})}{p(u_{i+t})}]\big)}\big]
    \end{flalign}
    Then, simplify the term $\mathop{\mathbb{E}}_{B_{negative}}[\frac{p(u_{i+t}\mid s_{i})}{p(u_{i+t})}]$. Since $\frac{p(u_{i+t}\mid s_{i})}{p(u_{i+t})}$ is the ratio of two continuous probability densities, it is also continuous and thus we can write the Expectation term in integral: 
    \begin{flalign}
    \mathop{\mathbb{E}}_{B}[\frac{p(u\mid s}{p(u)}] &= \int_{B} \frac{p(u\mid s)}{p(u)}p(u)du \\
                    &= \frac{1}{p(s)}\int_{B}\frac{p(u,s)}{p(u)}p(u)du \\
                    &= \frac{1}{p(s)}\int_{B} p(u,s)du \\    
                    &= \frac{1}{p(s)} p(s) \\
                    &= 1
    \end{flalign}
    Substitue $\mathop{\mathbb{E}}_{B}[\frac{p(u\mid s}{p(u)}]$ back in $\mathcal{L}^{opt}$ and we get: 
    \begin{flalign}
    \mathcal{L}^{opt} &= \mathop{\mathbb{E}}_{B} \mathop{\mathbb{E}}_{T}\big[\log{\big(1+\frac{p(u_{i+t})}{p(u_{i+t}\mid s_{i})}(N-1)\big)}\big]
    \end{flalign}
    In addition, since random variables $U$ and $S$ both are sampled from the sample distribution $P_{data}$, $P(U) \leq P(U\mid S)$ (the uncertainty of a random variable becomes smaller once another variable is fixed). Therefore we have the following relationship: 
    \begin{flalign}
    \mathcal{L}^{opt} &\geq \mathop{\mathbb{E}}_{B} \mathop{\mathbb{E}}_{T}\big[\log\big(\frac{p(u_{i+t}}{p(u_{i+t}\mid s_{i})}N\big)\big]\\
                    &= \mathop{\mathbb{E}}_{B} \mathop{\mathbb{E}}_{T}\big[\log\big(\frac{p(u_{i+t}}{p(u_{i+t}\mid s_{i})}\big)\big] + \mathop{\mathbb{E}}_{B} \mathop{\mathbb{E}}_{T}\big[\log N\big] \\
                    &= -\mathop{\mathbb{E}}_{B} \mathop{\mathbb{E}}_{T}\big[\log\big(\frac{p(u_{i+t}\mid s_{i})}{p(u_{i+t}}\big)\big] + \mathop{\mathbb{E}}_{B} \mathop{\mathbb{E}}_{T}\big[\log N\big] \\
                    &= -I(u_{i+t}; s_{i}) + \mathop{\mathbb{E}}_{B} \mathop{\mathbb{E}}_{T}\big[\log N\big]
    \end{flalign}
    Therefore, the lower bound for $I(u_{i+t}; s_{i})$ is: 
    \begin{flalign}
    I(u_{i+t}; s_{i}) \geq \mathop{\mathbb{E}}_{B} \mathop{\mathbb{E}}_{T}\big[\log N\big] - \mathcal{L}^{opt}
    \end{flalign}
    Minimizing the loss $\mathcal{L}$ will lead to maximizing the mutual information $I$.
    \end{proof}
    
    \subsection{Shared Encoder Approach}
    The original proposed CPC model contains only one autoregressive model - an unidirectional RNN. The unidirectional RNN context vectors from the first few frames of a speech signal can be inaccuracte since the RNN has only seen a few frames. It is therefore common to have a bidirectional RNN instead, such as for machine translation applications. However, similar to language modeling such as n-gram language model, the CPC model is trained on future frames prediction and birdirectional RNN, which takes in the whole sequence, contradicts our NCE training objective.  
    
    we took inspiration from \citep{peters2018deep}, which have two separate RNNs, one for forward sequence and one for backward sequence. The two RNNs are jointly trained, and the hidden states are later concatenated together for next word prediction. We proposed the shared encoder approach - two autoregressive models in the same latent space, illustrated in Figure \ref{fig:double_cpc_model_train}. Compare to the single autoregressive model, the shared encoder approach has an additinoal autoregressive model for the backward sequence. The two autoregressive models do frame predictions separately but are optimized jointly with the loss:
    \begin{flalign}
    \mathcal{L}_{joint} = -\frac{1}{2}\mathop{\mathbb{E}}_{B} \mathop{\mathbb{E}}_{T} \big[\log{\frac{f_{p1}(u_{i+t},s_{i})}{\sum_{B} f_{n1}(u_{i+t},s_{i})}} + \log{\frac{f_{p2}(u_{i+t},s_{i})}{\sum_{B} f_{n2}(u_{i+t},s_{i})}} \big],
    \end{flalign}
    where $f_{1}$ is the density ratio from the autoregressive model trained on forward sequence, and $f_{1}$ is the density ratio from the second autoregressive model trained on backward sequence. Similar to \citep{peters2018deep}, we concatenate the context vectors (hidden states) from the two autoregressive models during inference for downstream task (speaker verification). 
    
    \begin{knitrout}
    \definecolor{shadecolor}{rgb}{0.969, 0.969, 0.969}\color{fgcolor}\begin{figure}
    \centering\includegraphics[width=1.\maxwidth]{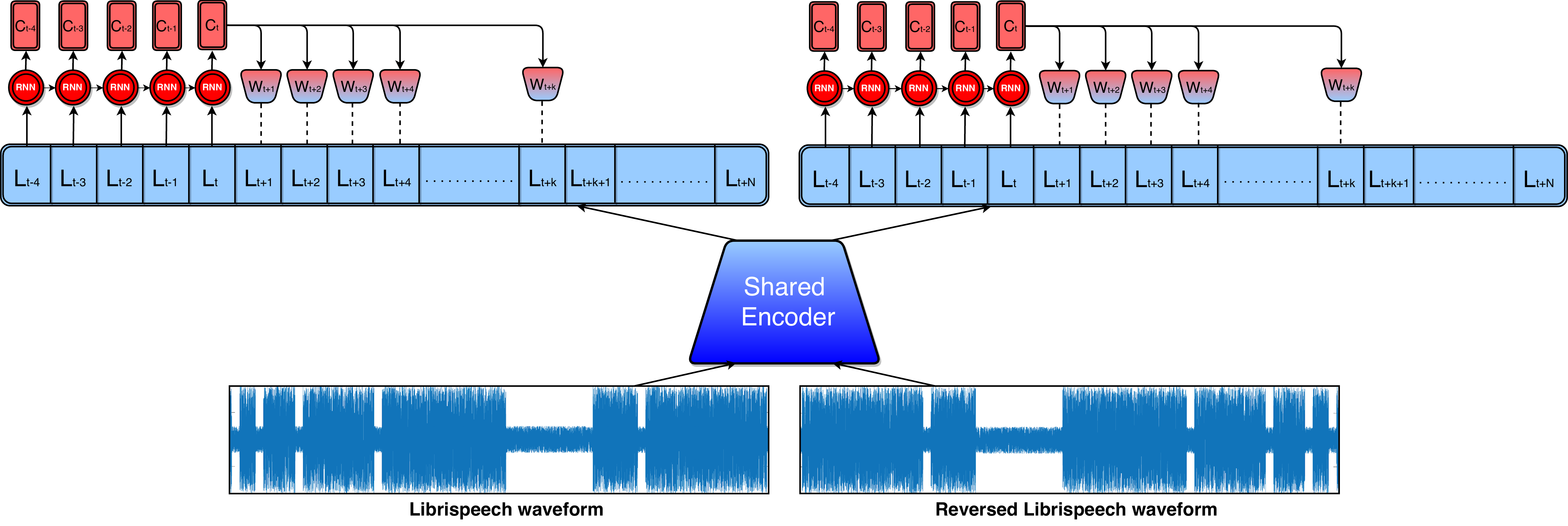} \caption[{\bf CPC Double Autoregressive Model} ]{{\bf CPC Double Autoregressive Model} Illustration of the CPC double autoregressive model's training stage. An waveform }\label{fig:double_cpc_model_train}
    \end{figure}
    \end{knitrout}
    
    \subsection{Detailed Implementation}
    Most of the CPC model implementation conforms to \citep{oord2018representation} with minor modifications. The raw waveform is input to the encoder without being processed with Voice Activity Detection or Mean Variance Normalization. In each training iteration, a segment of 1.28 seconds (or 20480 data points) is randomly extracted from the original waveform for every utterance, before inputting to the encoder. The encoder is a five layers 1-dimensional Convolutional Neural Network (CNN) with a 160 downsampling factor. For each of the five layers, the filter (kernel) sizes are $[10, 8, 4, 4, 4]$, the strides are $[5, 4, 2, 2, 2]$, and the zero paddings are $[3, 2, 1, 1, 1]$. All five layers have 512 hidden dimension. In \citep{oord2018representation}, the autoregressive model is implemented as a GRU with 256 hidden dimension, and the context vector (hidden state) is used as the CPC feature for downstream tasks. However for standard speaker verification systems, 256 input feature dimension would cost weeks to train and therefore it is impractical. We explored three CPC models with different GRU hidden dimension, and a comparison of the three CPC models are detailed in Figure \ref{table:cpc_model_summary}. CDCK2 and CDCK5 are variants of the single autoregressive model approach, while CDCK6 is based on the shared encoder approach. 
    
    \begin{table}[]
    \centering
    \begin{tabular}{|c|c|c|c|c|}
    \hline
    CPC model ID & \begin{tabular}[c]{@{}c@{}}number of \\ GRU(s)\end{tabular} & \begin{tabular}[c]{@{}c@{}}GRU \\ hidden dim\end{tabular} & \begin{tabular}[c]{@{}c@{}}number of \\ GRU layers\end{tabular} & \begin{tabular}[c]{@{}c@{}}CPC \\ feature dim\end{tabular} \\ \hline
    CDCK2 & 1 & 256 & 1 & 256 \\ \hline
    CDCK5 & 1 & 40 & 2 & 40 \\ \hline
    CDCK6 & 2 & 128 & 1 & 256 \\ \hline
    \end{tabular}
    \caption{\bf{CPC Model Summaries}}\label{table:cpc_model_summary}
    \end{table}
    
    To implement the NCE loss $\mathcal{L}$, we draw negative samples from different utterances excluding the current utterance. This can be conveniently implemented by selecting the other samples in the same batch as the negative samples. The advantage of such implementation is that the negative samples can be drawn in one batch of the forward pass. Finally, the timestep $k$ for future frame prediction is set to 12, and the batch size $B$ is set to 64 for all CPC models. Figure \ref{fig:cpc_implementation} is a visualization of the details of our CPC model implementation.
    
    \begin{knitrout}
    \definecolor{shadecolor}{rgb}{0.969, 0.969, 0.969}\color{fgcolor}\begin{figure}
    \centering\includegraphics[width=1.0\maxwidth]{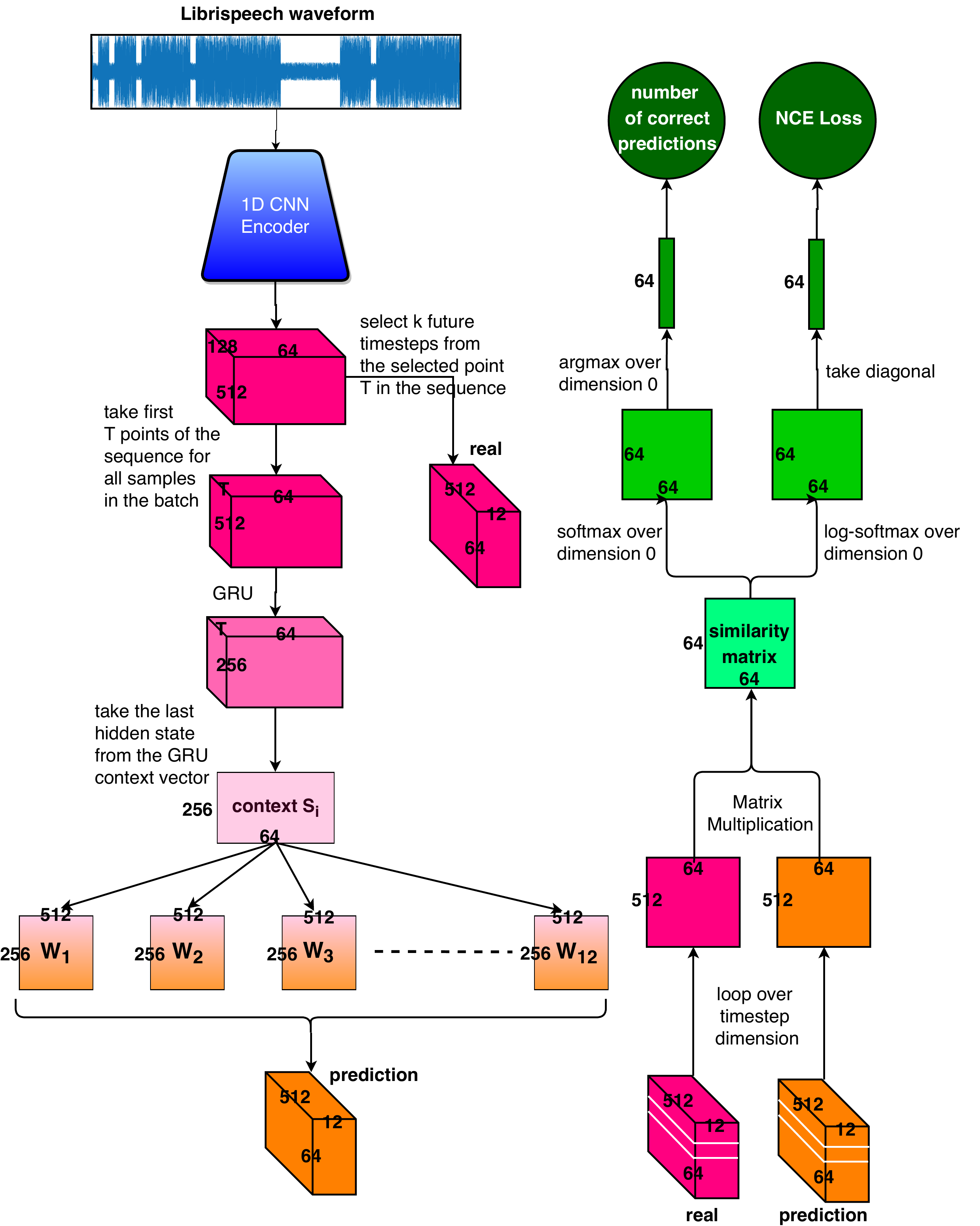} \caption[{\bf Implementation Details of CPC model} ]{{\bf Implementation Details of CPC model} Illustration of our CPC model implementation.}\label{fig:cpc_implementation}
    \end{figure}
    \end{knitrout}


\section{CPC-based Speaker Verification System}\label{sec:sec_cpc_ivector}
Since CPC feature learns high level information of the given input signal, it could contain relevant speaker information. We are interested in the effectiveness of the CPC feature in speaker verification, and how it fits in a standard speaker verification system. Figure \ref{fig:cpc_ivector_train} describes our CPC-based speaker verification system. The CPC model is trained on the training data, and frame-level representation is extracted by the model. To get a fixed-length utterance-level representation, we either temporally average across all frames for each utterance, or train an additional summarization system, the i-vector extractor. After getting the utterance-level representation, we first mean and length normalize across all representations, and train a Linear Discriminant Analysis to reduce feature dimension per utterance. Lastly, a decision generator, the PLDA model, is trained to get the log-likelihood ratio for each utterance before computing the EER. Figure \ref{fig:cpc_ivector_test} describe the testing pipeline for the CPC-based speaker verification system. 

    \begin{knitrout}
    \definecolor{shadecolor}{rgb}{0.969, 0.969, 0.969}\color{fgcolor}\begin{figure}
    \centering\includegraphics[width=1.0\maxwidth]{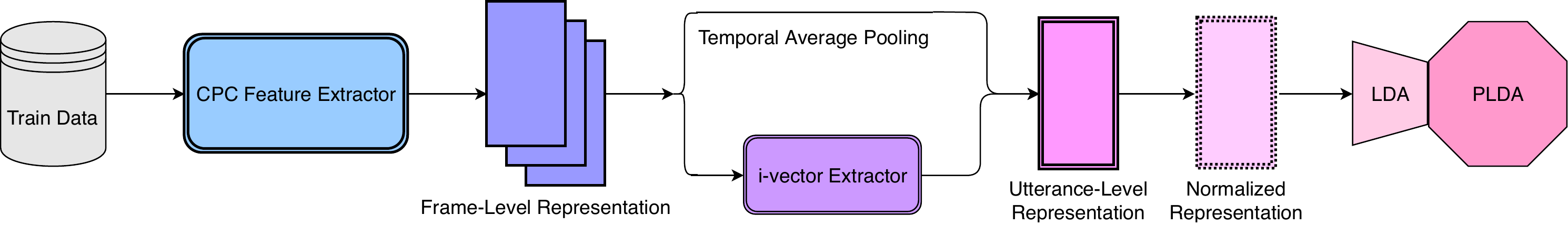} \caption[{\bf CPC-based Speaker Verification System - Training Pipeline} ]{{\bf CPC-based Speaker Verification System - Training Pipeline} Illustration of the training pipeline for CPC-based speaker verification system. }\label{fig:cpc_ivector_train}
    \end{figure}
    \end{knitrout}
    
    \begin{knitrout}
    \definecolor{shadecolor}{rgb}{0.969, 0.969, 0.969}\color{fgcolor}\begin{figure}
    \centering\includegraphics[width=1.0\maxwidth]{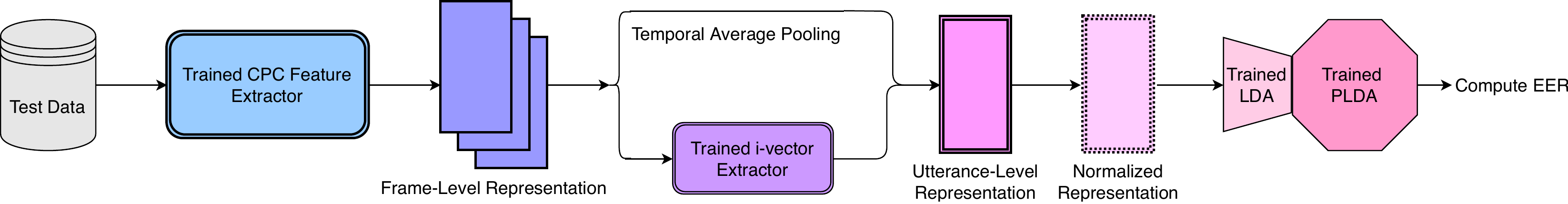} \caption[{\bf CPC-based Speaker Verification System - Testing Pipeline} ]{{\bf CPC-based Speaker Verification System - Testing Pipeline} Illustration of the testing pipeline for CPC-based speaker verification system. }\label{fig:cpc_ivector_test}
    \end{figure}
    \end{knitrout}


\cleardoublepage
\chapter{Experiments and Results}
\label{chap:experiments_results}

\section{LirbiSpeech}
We tested our CPC-model on the LibriSpeech corpus. LibriSpeech Corpus is an 1000-hour speech data set based on LibriVox's audio books \citep{panayotov2015librispeech}, and it consists of male and female speakers reading segments of book chapters. For example, 1320-122612-0000 means 'Segment 0000 of Chapter 122612 read by Speaker 1320.' The speech data is recorded at 16k Hz. LibriSpeech Corpus is partitioned into 7 subsets, and the description of each subset is summarized in Figure \ref{fig:librispeech_table}. In our experiments, we used train-clean-100, train-clean-360, and train-clean-500 subsets for training. Dev-other and dev-test are used as validation and CPC model selection. Finally, we report our speaker verification results on test-clean. 

    \begin{knitrout}
    \definecolor{shadecolor}{rgb}{0.969, 0.969, 0.969}\color{fgcolor}\begin{figure}
    \centering\includegraphics[width=0.8\maxwidth]{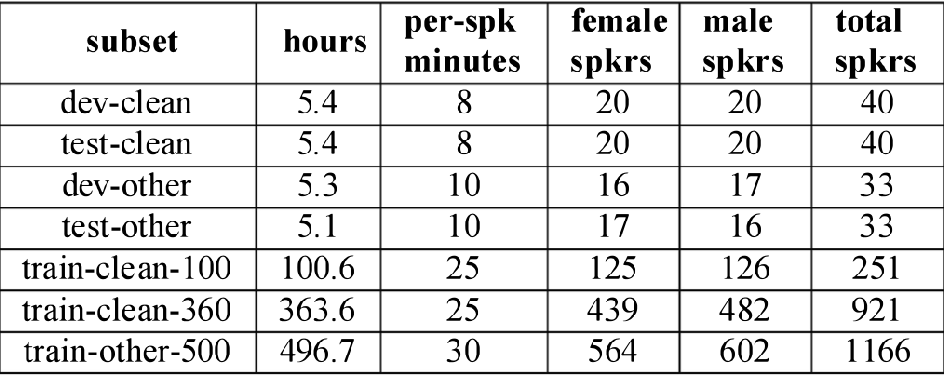} \caption[{\bf LibriSpeech Corpus Summary} ]{{\bf LibriSpeech Corpus Summary - number of hours and number of speakers} \citep{panayotov2015librispeech} }\label{fig:librispeech_table}
    \end{figure}
    \end{knitrout}

\section{Speaker Verification Trial List}
Since LibriSpeech is originally created for speech recognition, we have to manually create the speaker verification trial list. The trial list contains two three columns: enrollment ID, test ID and target/nontarget. The enrollment ID column contains the speech recordings that are enrolled, the test recordings are those tested against the enrollment recordings, and the target/nontarget indicates whether the speaker of the given test recording matches the speaker of the given enrollment recording. Table \ref{tab:trail_list_example} contains three example trials. 

\begin{table}[ht]
\centering
\begin{tabular}{|c|c|c|}
\hline
enrollment ID & test ID & target/nontarget \\ \hline
908-157963-0027 & 4970-29095-0029 & nontarget \\ \hline
908-157963-0027 & 908-157963-0028 & target \\ \hline
1320-122612-0007 & 4446-2275-0017 & nontarget \\ \hline
\end{tabular}
\caption{\bf{Example of Speaker Verification Trials}}\label{tab:trail_list_example}
\end{table}

We prepared our trial list in two different ways. The first trial list is created by randomly selecting half of the LibriSpeech recordings as enrollment and the other half as test. There are a total of 1716019 trials in the first trial list. The second trial list is also created in the same manner but we made sure that there is no overlap in chapters spoken by the same speaker. For example, the trial '1320-122617-0000 1320-122617-0025 target' is allowed in the first trial list but not in the second trial list. The two trial lists we described above are available for download: first trial list\footnote{\url{https://drive.google.com/open?id=10h9GH_vi-BRBT_L_xmSM1ZumQ__jRBmx}} and second trial list\footnote{\url{https://drive.google.com/open?id=1FDOU1iNSdGT-IMCQnuuJCWV421168x4H}}. 

\section{Speaker Verification EER}
We presented the model training results and speaker verification error rate of the three CPC models we implemented in Table \ref{table:cpc_model_train_summary}. CDCK2 and CDCK5 are trained for 60 iterations, and CDCK6 is trained for 30 iterations due to time limitation. CDCK5 has around 1.8 million less model parameters than CDCK2 and CDCK6 because its GRU hidden dimension is 40, which is significantly smaller. Expectedly, due to the larger model size, CDCK2 and CDCK6 has smaller NCE losses $\mathcal{L}$ and higher positive sample prediction accuracies than CDCK5. Furthermore, CDCK6 attains higher prediction accuracies with half the training iterations, which suggests that the shared encoder approach is more powerful than the single autoregressive model approach. 

\begin{table}[]
\centering
\begin{tabular}{|c|c|c|c|c|}
\hline
CPC model ID & number of  epoch & model size & \begin{tabular}[c]{@{}c@{}}dev NCE \\ loss\end{tabular} & \begin{tabular}[c]{@{}c@{}}dev \\ accuracy \end{tabular} \\ \hline
CDCK2 & 60 & 7.42M & 1.6427 & 26.42 \\ \hline
CDCK5 & 60 & 5.58M & 1.7818 & 22.48 \\ \hline
CDCK6 & 30 & 7.33M & 1.6484 & 28.24 \\ \hline
\end{tabular}
\caption{\bf{CPC Model Training Summaries}}\label{table:cpc_model_train_summary}
\end{table}

Figures \ref{fig:cdck2_accuracy}, \ref{fig:cdck5_accuracy}, \ref{fig:cdck6_accuracy} are the future frame positive sample prediction accuracies for CDCK2, CDCK5, and CDCK6 respectively. Figures \ref{fig:cdck2_loss}, \ref{fig:cdck5_loss}, \ref{fig:cdck6_loss} are the NCE losses for CDCK2, CDCK5, and CDCK6 respectively. The reported loss and accuracy are performed on the dev set, and we can see that the losses decrease while the prediction accuracies increase over training iterations. Note that the NCE loss $\mathcal{L}$ is averaged over all future prediction timesteps $1, 2, .., k$, and the prediction accuracy is calculated only on the last timestep $k$. In our implementation, $k$ is set to 12. Therefore, $\mathcal{L}$ is averaged over 12 timesteps, but the positive sample prediction accuracy is on the $12^{th}$ timestep only. 

    \begin{knitrout}
    \definecolor{shadecolor}{rgb}{0.969, 0.969, 0.969}\color{fgcolor}\begin{figure}
    \centering\includegraphics[width=0.75\maxwidth]{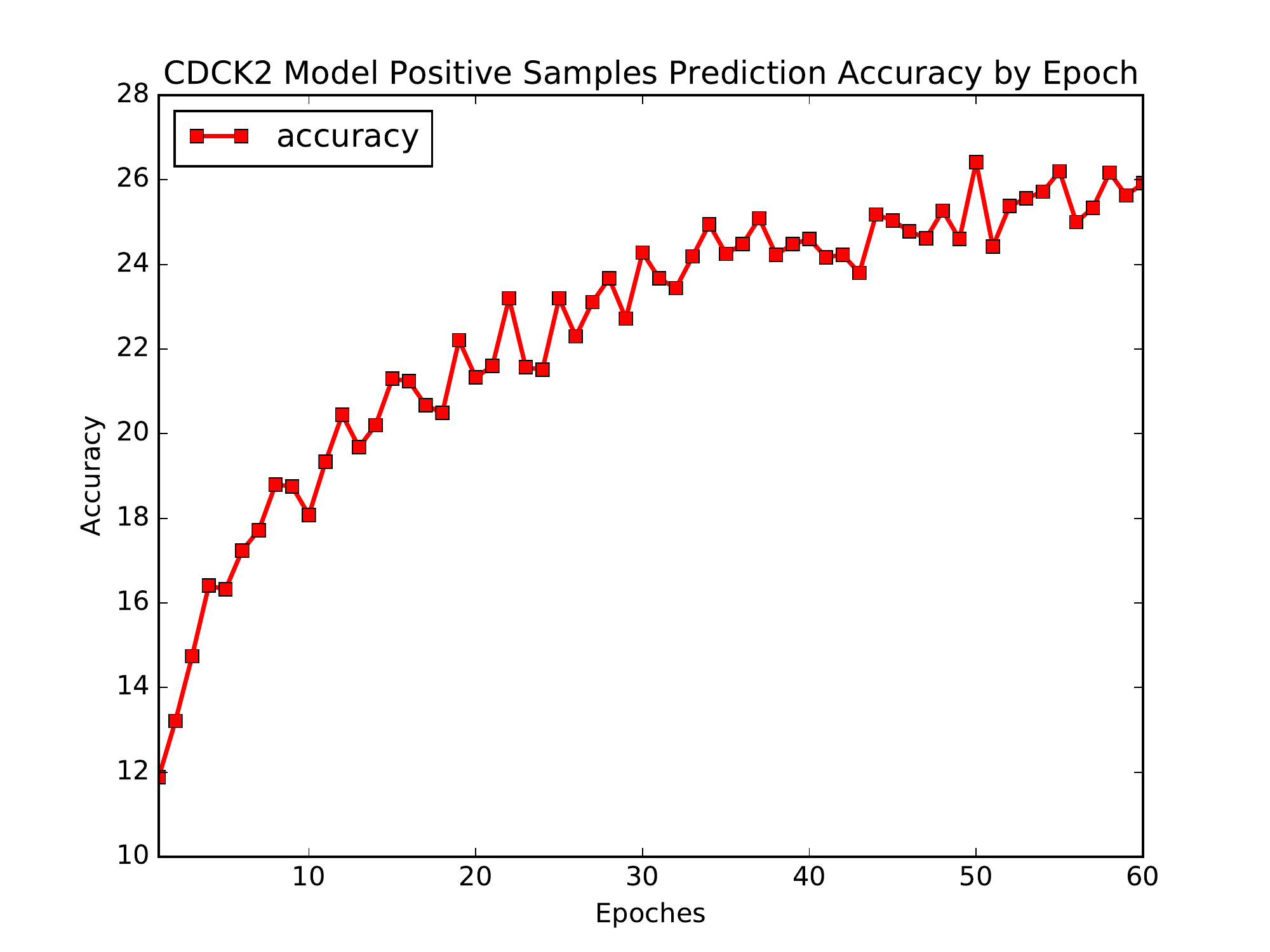} \caption[{\bf CDCK2 Model Positive Samples Prediction Accuracy of} ]{{\bf Positive Samples Prediction Accuracy of CDCK2 on development set over training iterations} }\label{fig:cdck2_accuracy}
    \end{figure}
    \end{knitrout}

    \begin{knitrout}
    \definecolor{shadecolor}{rgb}{0.969, 0.969, 0.969}\color{fgcolor}\begin{figure}
    \centering\includegraphics[width=0.75\maxwidth]{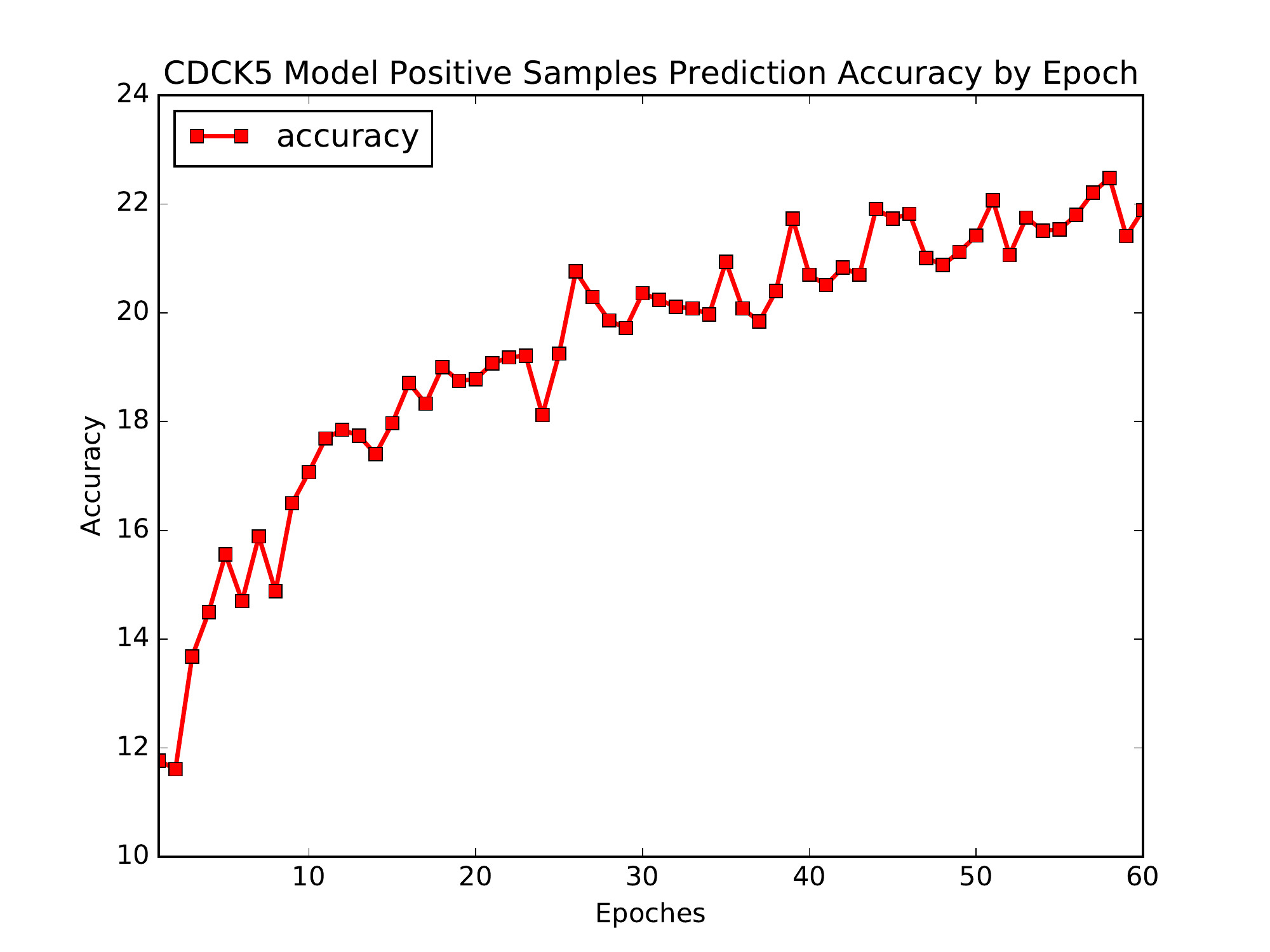} \caption[{\bf CDCK5 Model Positive Samples Prediction Accuracy} ]{{\bf Positive Samples Prediction Accuracy of CDCK5 on development set over training iterations} }\label{fig:cdck5_accuracy}
    \end{figure}
    \end{knitrout}
    
    \begin{knitrout}
    \definecolor{shadecolor}{rgb}{0.969, 0.969, 0.969}\color{fgcolor}\begin{figure}
    \centering\includegraphics[width=0.75\maxwidth]{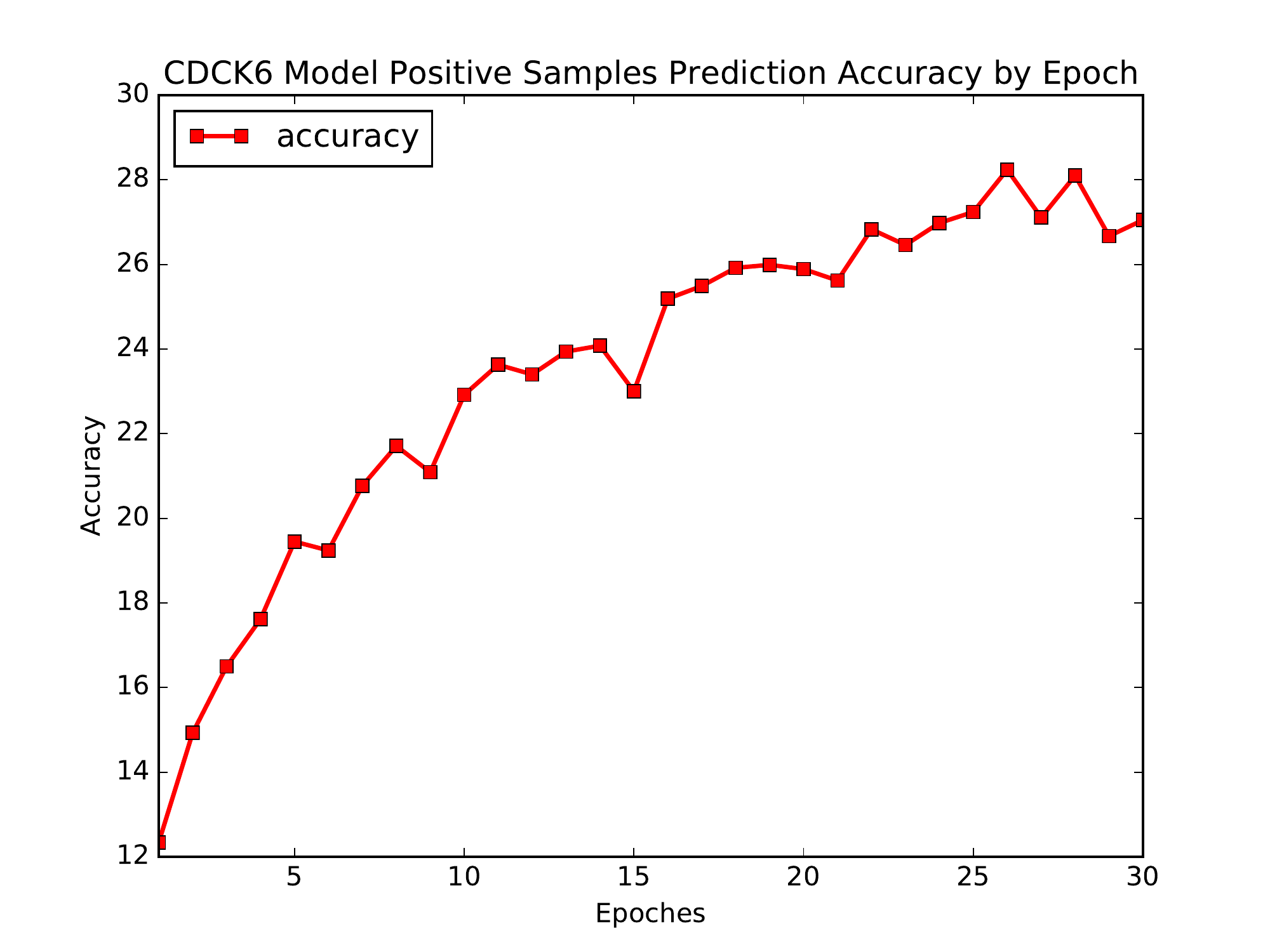} \caption[{\bf CDCK6 Model Positive Samples Prediction Accuracy} ]{{\bf Positive Samples Prediction Accuracy of CDCK6 on development set over training iterations} }\label{fig:cdck6_accuracy}
    \end{figure}
    \end{knitrout}

    \begin{knitrout}
    \definecolor{shadecolor}{rgb}{0.969, 0.969, 0.969}\color{fgcolor}\begin{figure}
    \centering\includegraphics[width=0.75\maxwidth]{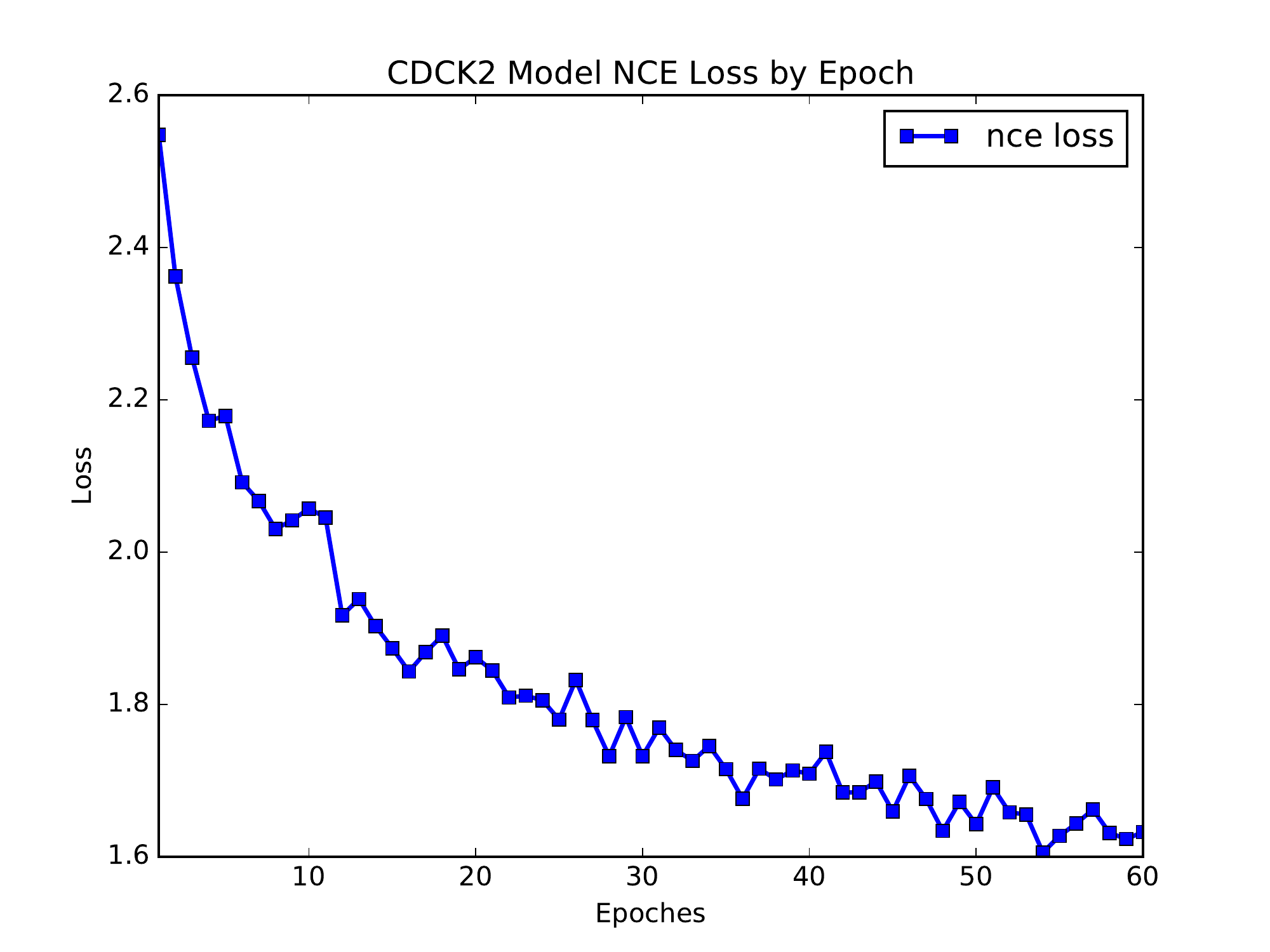} \caption[{\bf CDCK2 Model NCE Loss} ]{{\bf NCE Loss of CDCK2 on development set over training iterations} }\label{fig:cdck2_loss}
    \end{figure}
    \end{knitrout}

    \begin{knitrout}
    \definecolor{shadecolor}{rgb}{0.969, 0.969, 0.969}\color{fgcolor}\begin{figure}
    \centering\includegraphics[width=0.75\maxwidth]{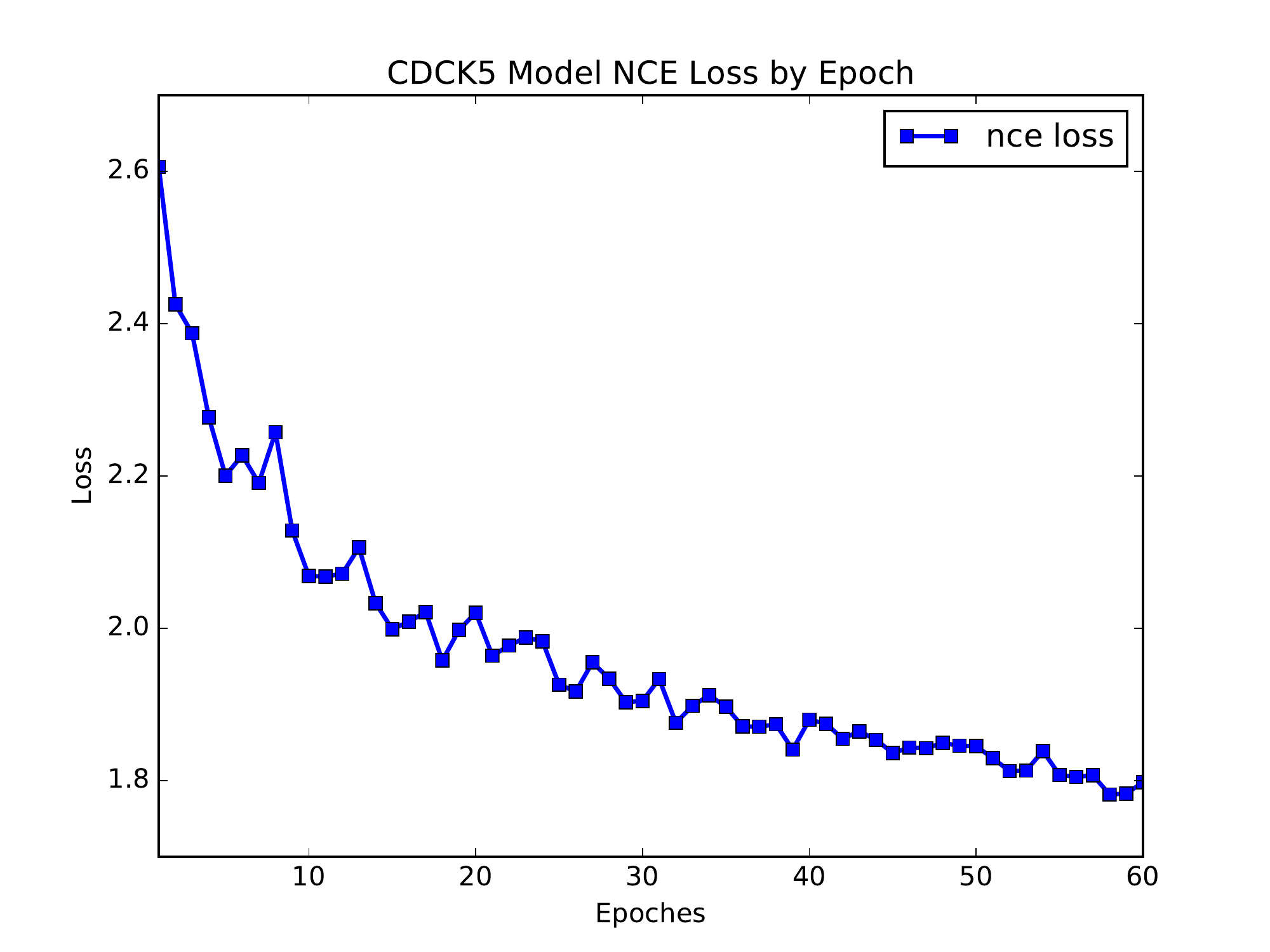} \caption[{\bf CDCK5 Model NCE Loss} ]{{\bf NCE Loss of CDCK5 on development set over training iterations} }\label{fig:cdck5_loss}
    \end{figure}
    \end{knitrout}
    
    \begin{knitrout}
    \definecolor{shadecolor}{rgb}{0.969, 0.969, 0.969}\color{fgcolor}\begin{figure}
    \centering\includegraphics[width=0.75\maxwidth]{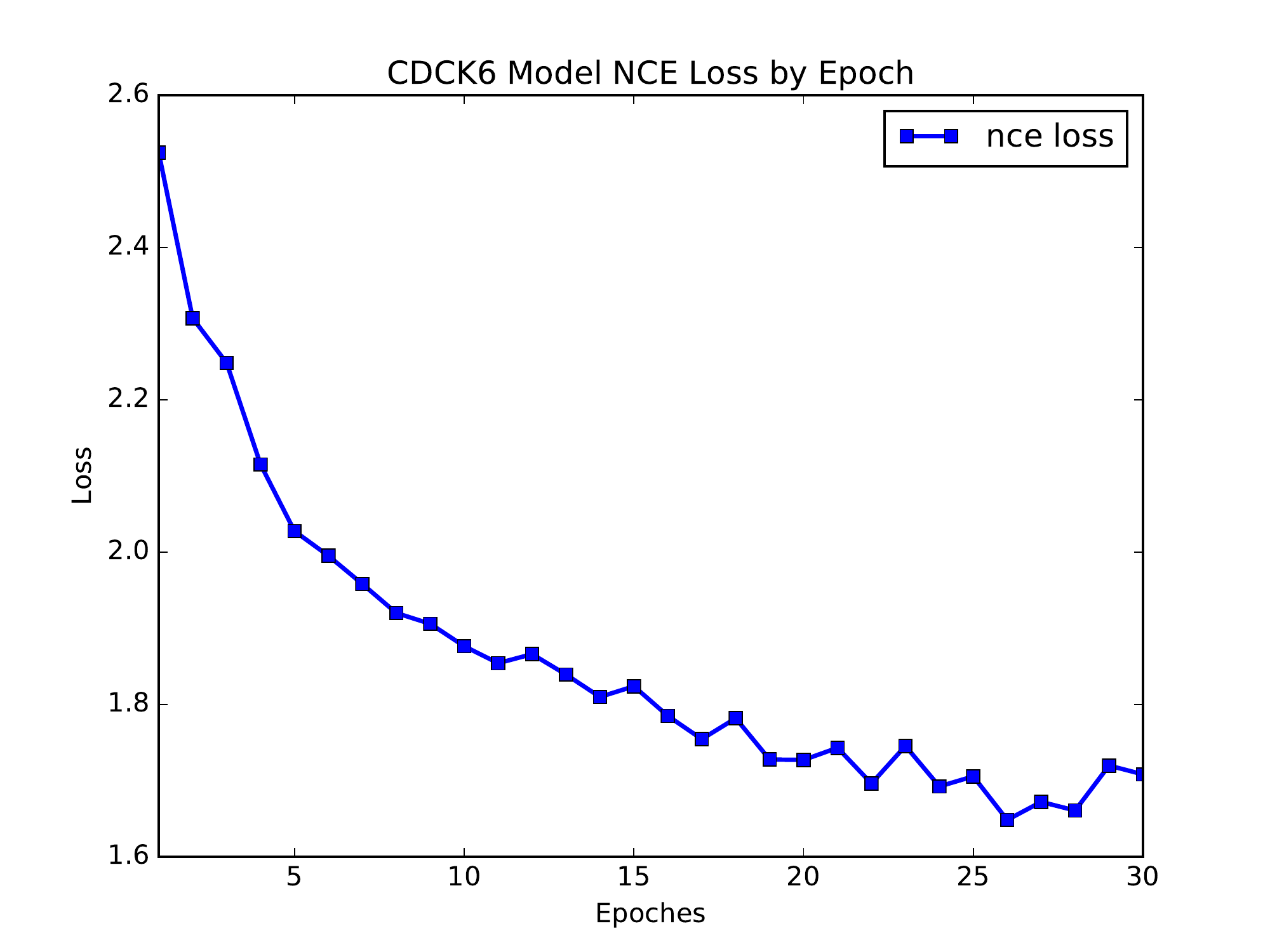} \caption[{\bf CDCK6 Model NCE Loss} ]{{\bf NCE Loss of CDCK6 on development set over training iterations} }\label{fig:cdck6_loss}
    \end{figure}
    \end{knitrout}

After the CPC models are trained, the context vectors (hidden states) of the models are extracted as the CPC features. These features are used as the input feature for speaker verification. We explored two approaches to summarization in the speaker verification system described in Figure \ref{fig:cpc_ivector_train}: temporal average pooling and i-vectors. In the first approach, temporal average pooling, frame-level features are averaged across frames to get a fixed-length utterance-level feature for each utterance. The speaker verification results of the CPC features and the baseline MFCC features with temporal average pooling is summarized in Table \ref{table:speaker_verification_a}. We can first see that the speaker verification EER of the first trial list is significantly lower than that of the second trial list. This is expected since the second trial list contains no speaker-chapter overlap between enrollment and test, and thus the higher error rate. Secondly, CPC features show significant improvement over MFCC. Specifically, features from CDCK2 model recorded $5.887$ and $13.48$ EER, which are $36\%$ and $18\%$ relative improvements over the baseline. Although CDCK6 showed lower NCE loss and higher prediction accuracies during training, its features performed worse than the ones from CDCK2. 

\begin{table}[]
\centering
\begin{tabular}{|c|c|c|c|c|c|}
\hline
Feature & Feature Dim & Summarization & LDA Dim & 1st EER & 2nd EER \\ \hline
MFCC & 24 & average pooling & 24 & 9.211 & 13.48 \\ \hline
CDCK2 & 256 & average pooling & 200 & 5.887 & 11.1 \\ \hline
CDCK5 & 40 & average pooling & 40 & 7.508 & 12.25 \\ \hline
CDCK6 & 256 & average pooling & 200 & 6.809 & 12.73 \\ \hline
\end{tabular}
\caption{\bf{Speaker Verification Results on LibriSpeech test-clean-100 - Summarization with Average Pooling}}\label{table:speaker_verification_a}
\end{table}

The second approach to summarization in speaker verification is i-vectors, which also gives a fix-length utterance level feature for each utterance. However, as mentioned earlier, usually the feature dimension to i-vectors is below 60. A feature dimension of 256 will take weeks to train an i-vector extractor. Therefore, dimension reduction on frame-level CPC features is first performed before summarization. We chose Principal Componenet Anaysis (PCA) for reducing the CPC feature dimension because we do not want to introduce extra nonlinearity for the learned feature and PCA is a linear transform. Table \ref{table:cpc_pca} is the summary of the CPC features after PCA transform with their corresponding PCA variance ratio, and the feature dimensions are all smaller or equal to 60 after PCA. 

\begin{table}[]
\centering
\begin{tabular}{|c|c|c|c|}
\hline
Feature w PCA & Original Feature & PCA Dim & PCA Variance Ratio \\ \hline
CDCK2-36 & CDCK2 & 36 & 76.76 \\ \hline
CDCK2-60 & CDCK2 & 60 & 87.40 \\ \hline
CDCK5-24 & CDCK5 & 24 & 93.39 \\ \hline
CDCK6-36 & CDCK6 & 36 & 82.30 \\ \hline
CDCK6-60 & CDCK6 & 60 & 90.31 \\ \hline
\end{tabular}
\caption{\bf{CPC features applied with PCA Summary}}\label{table:cpc_pca}
\end{table}

Table \ref{table:speaker_verification_b} presents the result of various MFCC, CPC, and combinations of MFCC and CPC features for speaker verificaiton with i-vectors. We can see that i-vectors with MFCC alone got $5.518$ and $8.157$ EER on the two trial lists. We trained three i-vectors systems with CPC features after PCA: CDCK2-60, CDCK5-24, and CDCK6-60. We can see that these features achieved up to $11\%$ EER relative improvement over the baseline on the first trial list. The relative improvements are much smaller compare to their counterparts in Table \ref{table:speaker_verification_a}. Furthermore, on the second trial list, MFCC with i-vectors prevails CPC with i-vectors. 

\begin{table}[]
\centering
\begin{tabular}{|c|c|c|c|c|}
\hline
Feature & Feature Dim & Summarization & 1st EER & 2nd EER \\ \hline
MFCC & 24 & i-vectors & 5.518 & 8.157 \\ \hline
CDCK2-60 & 60 & i-vectors & 5.351 & 9.753 \\ \hline
CDCK5-24 & 24 & i-vectors & 4.911 & 8.901 \\ \hline
CDCK6-60 & 60 & i-vectors & 5.228 & 9.009 \\ \hline
MFCC + CDCK2-36 & 60 & i-vectors & 3.62 & 6.898 \\ \hline
MFCC + CDCK5-24 & 48 & i-vectors & 3.712 & 6.962 \\ \hline
MFCC + CDCK6-36 & 60 & i-vectors & 3.691 & 6.765 \\ \hline
\end{tabular}
\caption{\bf{Speaker Verification Results on LibriSpeech test-clean-100 - Summarization with i-vectors}}\label{table:speaker_verification_b}
\end{table}

Since MFCC and CPC are two very different feature extraction methods, they should capture different aspects of the speech signal, which may be complementary for speaker verification. We fused MFCC and CPC features before i-vectors by simply concatenating the two feature vectors. The last three rows of Table \ref{table:speaker_verification_b} show the results of fusing MFCC with CPC features after PCA. We can see that the best combinations attains $34\%$ and $17\%$ relative improvements over MFCC i-vectors on the two lists. 

\section{Feature Visualizations}

\begin{figure}[tb]
 \centering
 \includegraphics[width=\columnwidth]{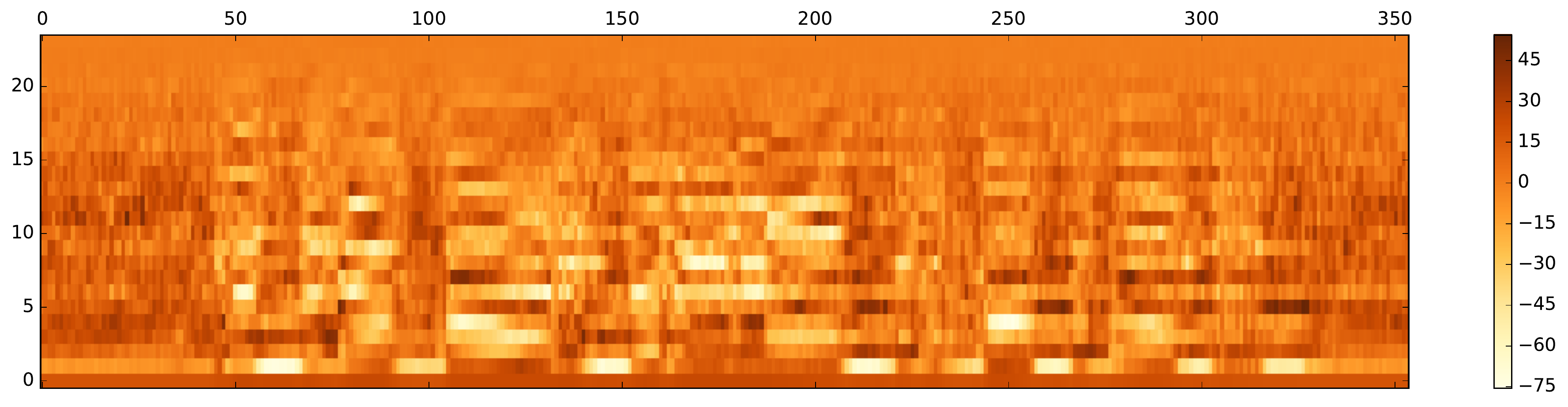}
 \includegraphics[width=\columnwidth]{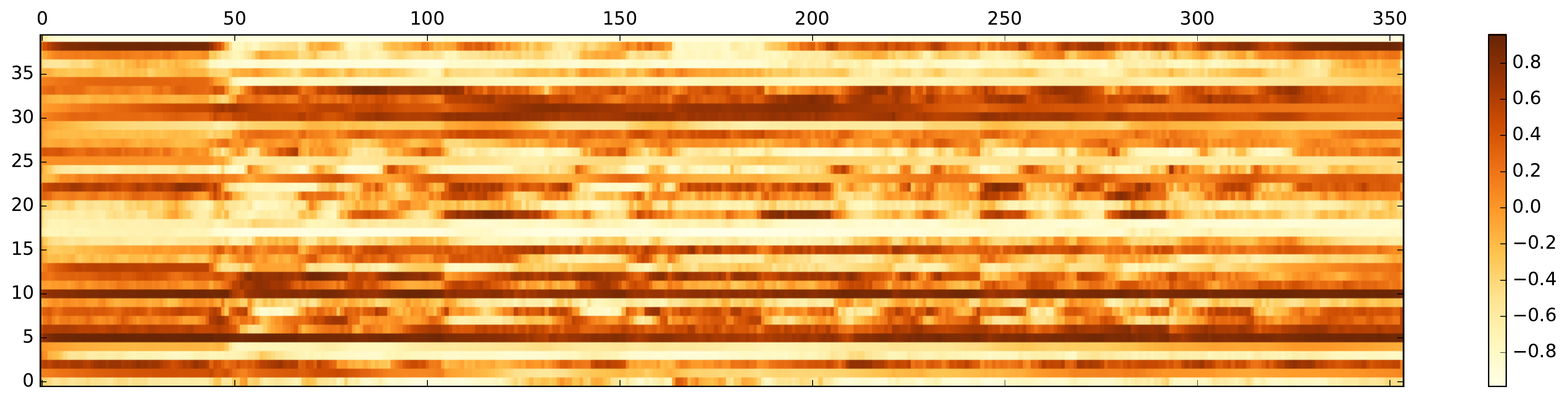}
 \includegraphics[width=\columnwidth]{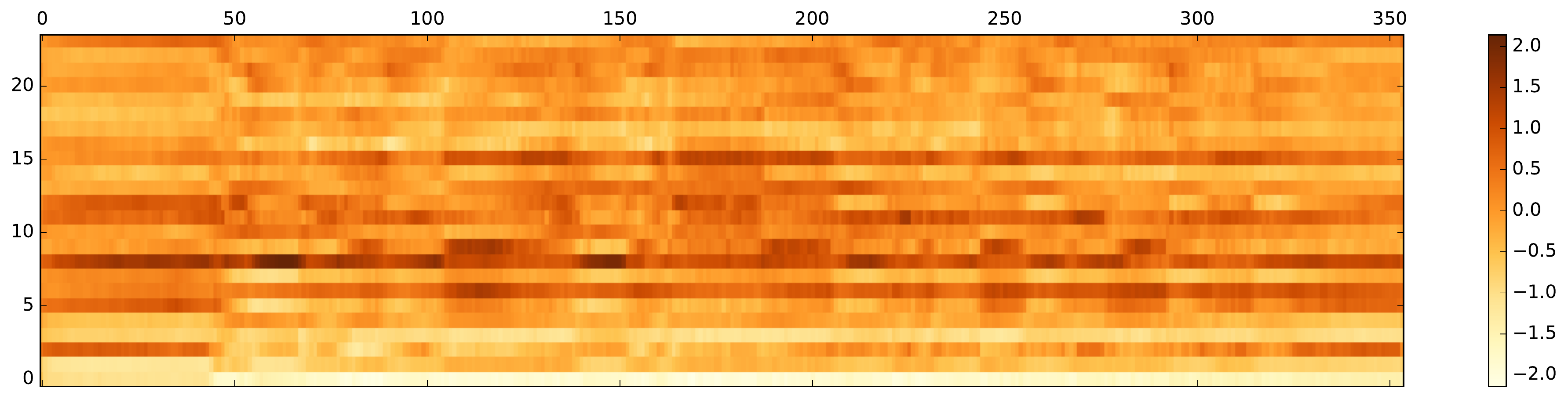}
\caption[{\bf Feature Visualization for Utterance 2830-3980-0028} ]{{\bf A visual comparison of MFCC (top), CPC (middle), and CPC with PCA (bottom) for utterance 2830-3980-0028}}
\label{fig:visualize_2830-3980-0028}
\end{figure}

\begin{figure}[tb]
 \centering
 \includegraphics[width=\columnwidth]{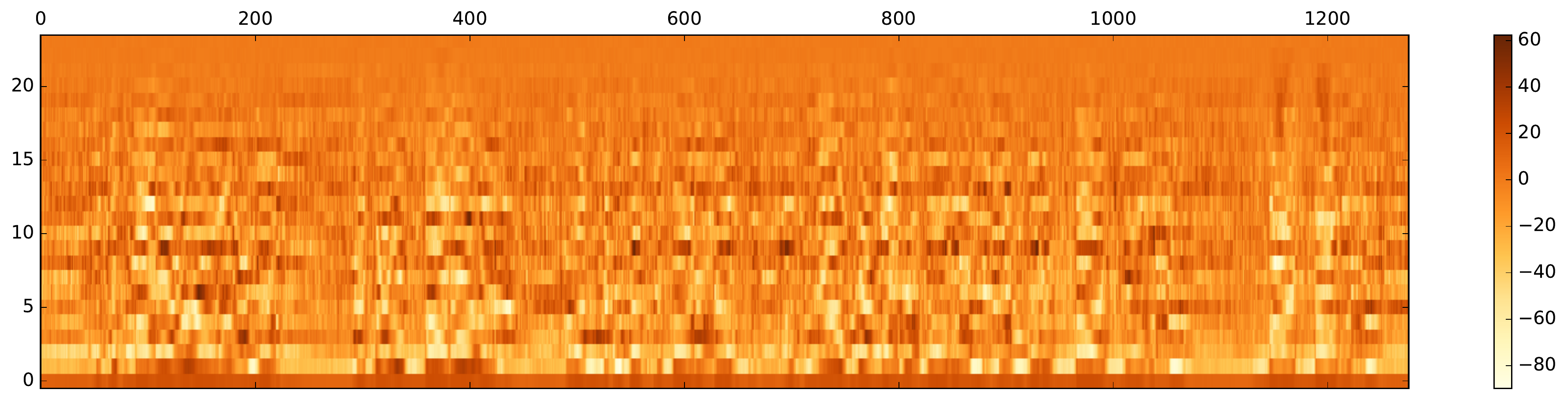}
 \includegraphics[width=\columnwidth]{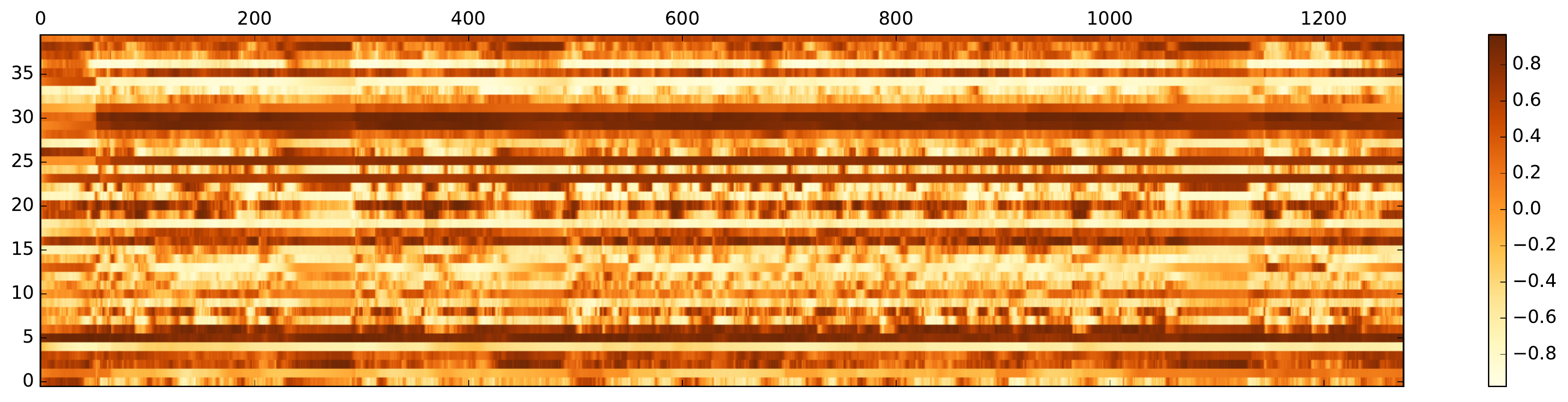}
 \includegraphics[width=\columnwidth]{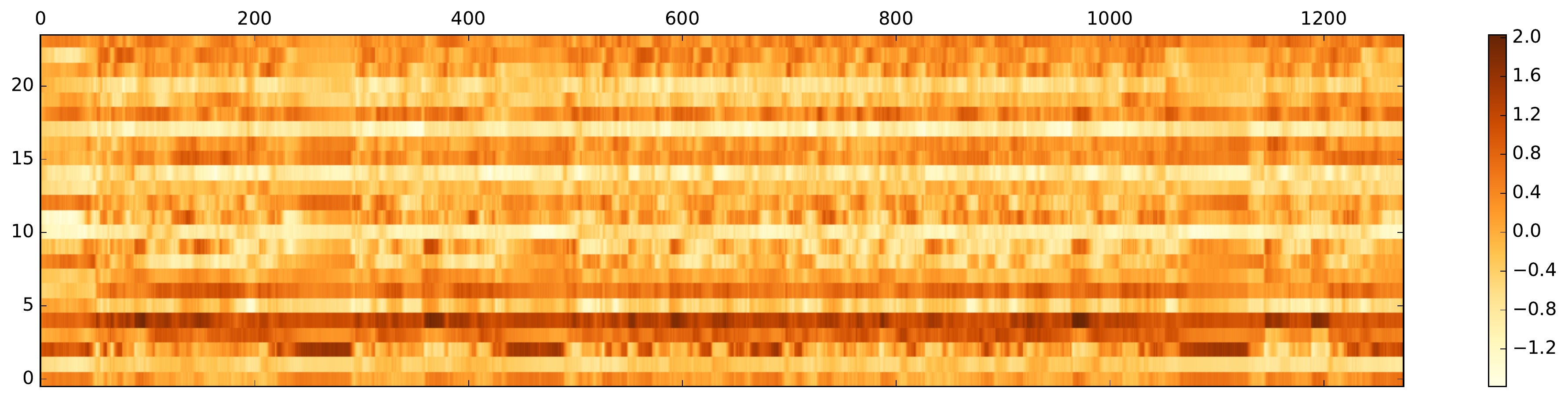}
\caption[{\bf Feature Visualization for Utterance 5105-28241-0017} ]{{\bf A visual comparison of MFCC (top), CPC (middle), and CPC with PCA (bottom) for utterance 5105-28241-0017}}
\label{fig:visualize_5105-28241-0017}
\end{figure}

It is a good practice to visualize speech features, and we visualize the CPC features and compare them to MFCC. Since CPC features from model CDCK2 and CDCK6 are 256 dimension, which may contain too much visual details, we chose to visualize CPC feature from CDCK5, which has 40 dimension. Figure \ref{fig:visualize_2830-3980-0028} and \ref{fig:visualize_5105-28241-0017} are visual comparisons of MFCC and CPC features on two randomly picked LibriSpeech test-clean-100 utterances: 2830-3980-0028 and 5105-28241-0017. We also visualize CPC features with PCA transform, CDCK5-24. Looking at the visualizations, CPC and MFCC bear very little similarity that they differ in structure and magnitude. However, one observation worth noting of the CPC features is that there are several feature bins whose values remain in a small range over time, which signifies that the CPC features learn some global information that lasts over time. 

\section{Speaker Verificaiton DET Curves}
To examine the tradeoff between false alram and miss rate, we plotted the Detection Error Tradeoff (DET) curves for the CPC and MFCC based speaker verification system. Figure \ref{fig:det_curve_1_trial_1} and \ref{fig:det_curve_1_trial_2} are DET curves for MFCC and CPC fusion-based i-vectors speaker verification system. For both trial lists, we can see that the fusion features reduced the miss and false alarm probabilities compared to the baseline. 

    \begin{knitrout}
    \definecolor{shadecolor}{rgb}{0.969, 0.969, 0.969}\color{fgcolor}\begin{figure}
    \centering\includegraphics[width=.8\maxwidth]{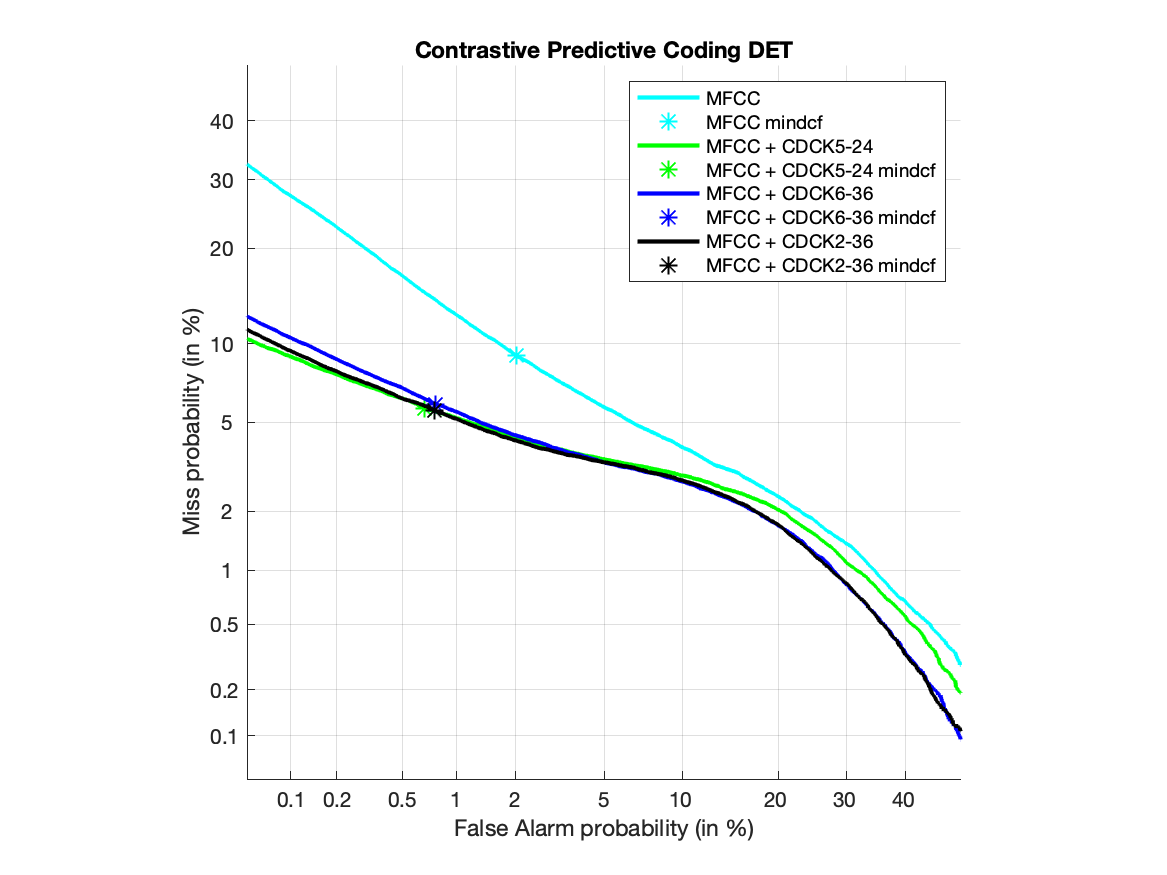} \caption[{\bf 1st Trial List DET Curve for CPC and MFCC Fusion i-vectors Speaker Verification System}]{{\bf 1st trial list DET curve for CPC and MFCC feature-level fusion i-vectors speaker verification system} }\label{fig:det_curve_1_trial_1}
    \end{figure}
    \end{knitrout}
    
    \begin{knitrout}
    \definecolor{shadecolor}{rgb}{0.969, 0.969, 0.969}\color{fgcolor}\begin{figure}
    \centering\includegraphics[width=.8\maxwidth]{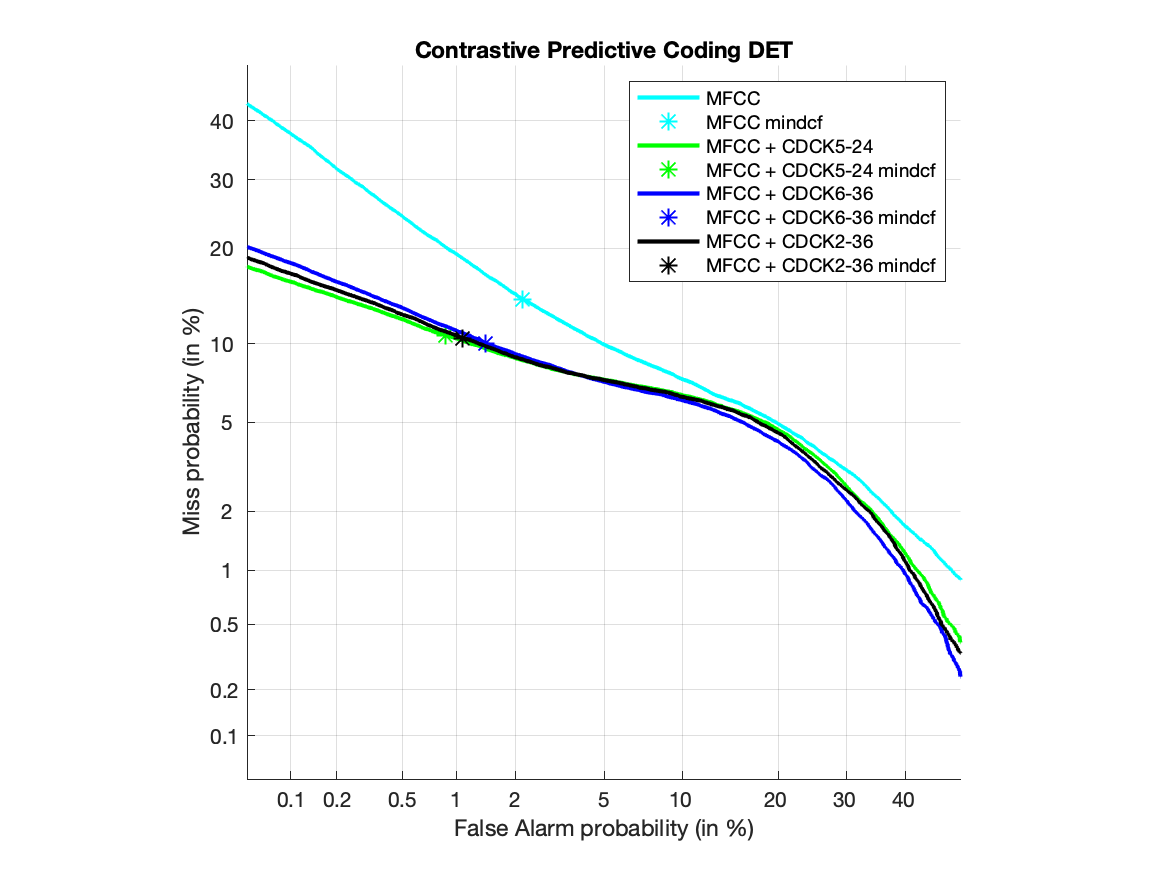} \caption[{\bf 2nd Trial List DET Curve for CPC and MFCC Fusion i-vectors Speaker Verification System} ]{{\bf 2nd trial list DET curve for CPC and MFCC feature-level fusion i-vectors speaker verification system} }\label{fig:det_curve_1_trial_2}
    \end{figure}
    \end{knitrout}
    
    Figure \ref{fig:det_curve_2_trial_1} and \ref{fig:det_curve_2_trial_2} are DET curves for CPC i-vectors speaker verification system. CPC i-vectors attained lower miss and false alarm probabilities only on the first trial list compared to MFCC i-vectors. 
    
    \begin{knitrout}
    \definecolor{shadecolor}{rgb}{0.969, 0.969, 0.969}\color{fgcolor}\begin{figure}
    \centering\includegraphics[width=.8\maxwidth]{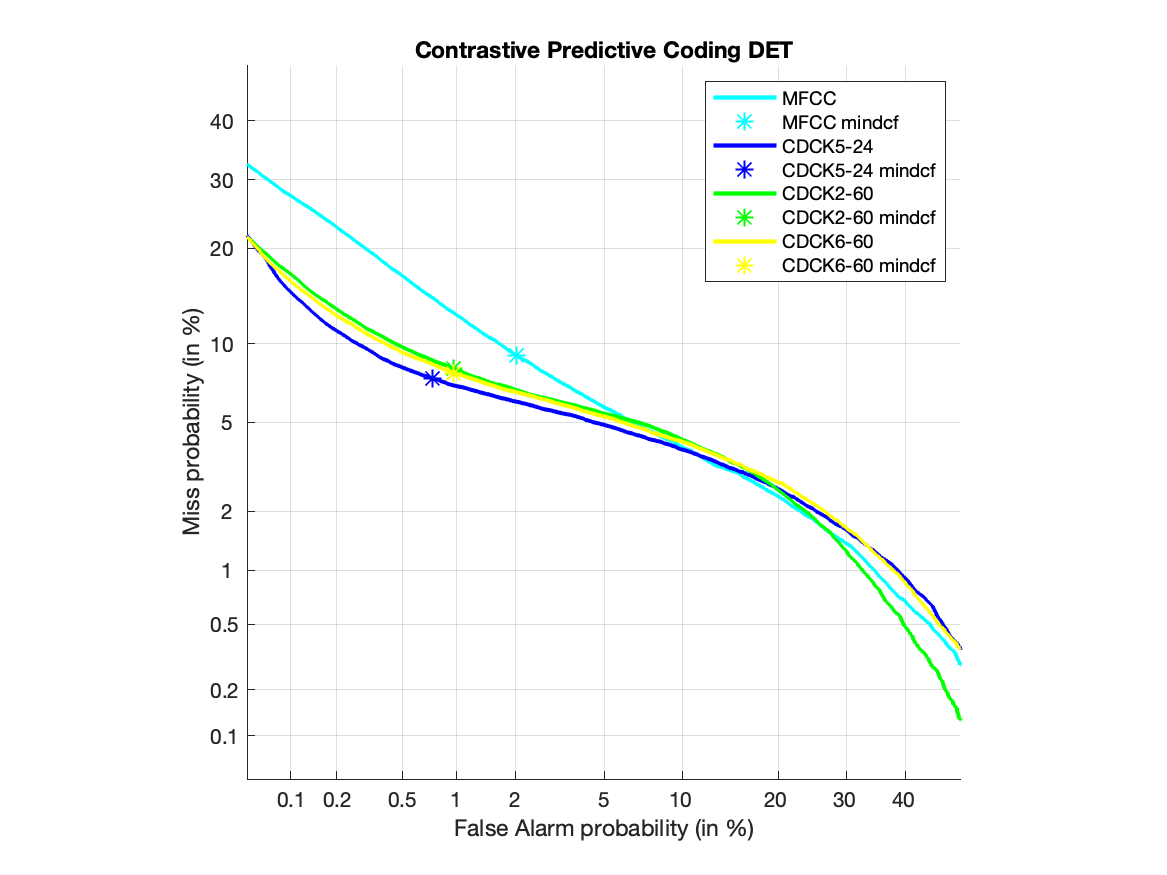} \caption[{\bf 1st Trial List DET Curve for CPC i-vectors Speaker Verification System} ]{{\bf 1st trial list DET curve for CPC i-vectors speaker verification system} }\label{fig:det_curve_2_trial_1}
    \end{figure}
    \end{knitrout}
    
    \begin{knitrout}
    \definecolor{shadecolor}{rgb}{0.969, 0.969, 0.969}\color{fgcolor}\begin{figure}
    \centering\includegraphics[width=.8\maxwidth]{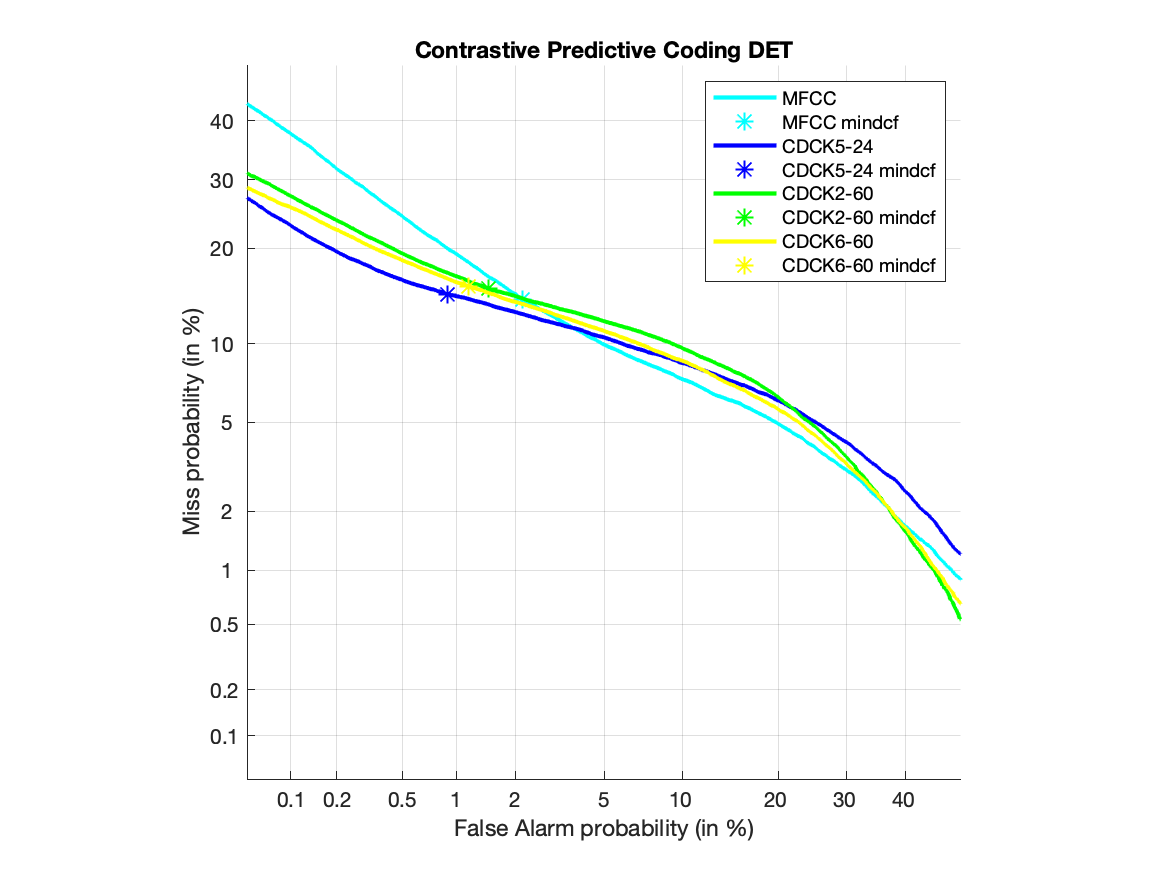} \caption[{\bf 2nd Trial List DET Curve for CPC i-vectors Speaker Verification System} ]{{\bf 2nd trial list DET curve for CPC i-vectors speaker verification system} }\label{fig:det_curve_2_trial_2}
    \end{figure}
    \end{knitrout}
\cleardoublepage
\chapter{Discussion and Conclusion}
\label{chap:conclusion}

\section{CPC as an Alternative Feature for Speaker Verification}
Common speech and speaker recognition systems employed deterministic Fourier-Transform-based features, such as MFCC, FilterBanks, or Peceptual Linear Predictive (PLP). In this work, we explored an unsupervised learned feature, CPC, for speaker verification task. We showed that CPC attains competitive speaker verification accuracy on LibriSpeech corpus, and it is presented as a potential alternative feature for future speaker verification research.

\section{i-vectors is not an Ideal Summarization Method for CPC}
i-vectors is one of the most popular features for speech analysis tasks. It is widely used for speaker recognition, language identification, speech recognition, etc. However, one constraint that i-vectors imposed on the input feature is that it has multi-Guassian distributed. If the input feature does not comply to a multi-Guassian distribution, GMM-UBM and hence i-vectors would not likely to work. From our experiments, we observed that i-vectors is not an ideal summarization method, that summarizes frame-level feature into utterance-level feature, for CPC compared to MFCC i-vectors. Compare Table \ref{table:speaker_verification_a}, which shows speaker verification EER of CPC features with average pooling, and Table \ref{table:speaker_verification_b}, which shows the EER of CPC features with i-vectors. CPC shows very strong results over MFCC with average pooling as the summarization method. On the other hand, when i-vectors is used as the summarization method, CPC does not show clear advantage oer MFCC. One speculation is that CPC features are not multi-Guassian distributed, and hence there may be better summarization method, such as the x-vectors, which does not assume any input distribution on the input features. 

\section{CPC Complements MFCC for i-vectors Speaker Verification}
We observed that CPC complements MFCC for i-vectors speaker verification system. Table \ref{table:speaker_verification_b} contains results of CPC and MFCC feature fusion with i-vectors, which give improvements over both MFCC i-vectors and CPC i-vectors. Similarly, Figure \ref{fig:det_curve_1_trial_1} and \ref{fig:det_curve_1_trial_2} are the fusion i-vectors DET curves, which are better than that of CPC features \ref{fig:det_curve_2_trial_1} and \ref{fig:det_curve_2_trial_2}. Therefore, we hypothesize that CPC complements MFCC for i-vectors based speaker verification system on the LibriSpeech corpus. However, whether this is true for all speech data is left for future work. 

\section{Future Work}
    Looking ahead, there are several directions for this work worth exploring. We listed five potential improvements and applications we would like to work on in the near future.  
    
    \subsection{Density Estimation Methods}
    First of all, we followed \citep{oord2018representation} and used Noice Contrastive Estimation for estimating the density ratio for learning high-level representation. There are other possible density estimation methods we can experimented with, such as the Importance Sampling. We are curious with the effectiveness of NCE and how it compares to other density estimation methods. 
    
    \subsection{SRE16}
    Librispeech corpus is a relatively clean (little noise) datasets that was originally made for speech recognition. Although the results we presented show potentials, we have to tested on publicly recognized datasets. In addition, we manually created our own trial lists since LibriSpeech does not provide one. We could not compare our findings to other speaker verification systems. We are planning to conduct CPC model refinements and speaker verification experiments on NIST SRE16 with the data in Table \ref{table:sre16_train_list}\footnote{Based on \url{https://github.com/kaldi-asr/kaldi/tree/master/egs/sre16/v1}}.
    
    \begin{table}[]
    \centering
    \begin{tabular}{|c|c|}
    \hline
    Corpus & LDC Catalog No. \\ \hline
    SWBD2 Phase 1 & LDC98S75 \\ \hline
    SWBD2 Phase 2 & LDC99S79 \\ \hline
    SWBD2 Phase 3 & LDC2002S06 \\ \hline
    SWBD Cellular 1 & LDC2001S13 \\ \hline
    SWBD Cellular 2 & LDC2004S07 \\ \hline
    SRE2004 & LDC2006S44 \\ \hline
    SRE2005 Train & LDC2011S01 \\ \hline
    SRE2005 Test & LDC2011S04 \\ \hline
    SRE2006 Train & LDC2011S09 \\ \hline
    SRE2006 Test 1 & LDC2011S10 \\ \hline
    SRE2006 Test 2 & LDC2012S01 \\ \hline
    SRE2008 Train & LDC2011S05 \\ \hline
    SRE2008 Test & LDC2011S08 \\ \hline
    SRE2010 Eval & LDC2017S06 \\ \hline
    Mixer 6 & LDC2013S03 \\ \hline
    \end{tabular}
    \caption{\bf{Training Data List for SRE16}}\label{table:sre16_train_list}
    \end{table}
    
    \subsection{CPC x-vectors}
    As mentioned previously, i-vectors may not be the ideal summarization methods for CPC. We plan to conduct x-vectors \citep{snyder2018x} speaker verification experiments after switching to SRE16.
    
    \subsection{Language Identification}
    We would also like to conduct CPC experiments on language identification\footnote{Based on \url{https://github.com/kaldi-asr/kaldi/tree/master/egs/lre07}}., which uses techniques from speaker recognition. Since CPC is designed to capture global information, it should learn some degree of language information in addition to speaker information of a speech signal. 
    
    \subsection{Domain Adaptation for Speaker Recognition}
    Finally, we would like to apply CPC for speaker recognition domain adaptation. Although there are signs that CPC may not generalize well to unseen conditions \ref{table:speaker_verification_b}, we are interested to see how CPC can be used in that context. 

\cleardoublepage
\printbibliography[title={Bibliography}]
\end{refsection}
\chapter*{Vita}

Cheng-I Jeff Lai grew up in Taiwan. At age 15, he left home and rent a room at Taipei to study at Taipei Municipal Jianguo High School. At age 18, Cheng-I attended Johns Hopkins University with a desire to study biophysics until he met Prof. Najim Dehak, who convinced him the beauty and delicacy of human spoken language. He subsequently dedicated a good amount of his time on speech processing and speaker recognition research, with a focus on deep learning approahces to speech. In Cheng-I's Sophomore and Junior year, he interned at the Human Language Technology Center of Excellence (HLTCoE) and the Informatics Forum, University of Edinburgh. He will receive a Bachelor's degree in Electrical Engineering in December, 2018. Beginning February, 2019, Cheng-I will start as a research assistant at Center for Language and Speech Processing and also interview for Ph.D. programs. 

\end{document}